\RequirePackage{fix-cm}
\documentclass[smallextended]{svjour3}       
\smartqed  
\usepackage{graphicx}
\usepackage{acronym}
\usepackage{subfigure}
\usepackage[table]{xcolor}
\usepackage{multicol}
\usepackage{multirow}
\usepackage[export]{adjustbox}
\usepackage{amsmath}
\usepackage{algorithm}
\usepackage[noend]{algpseudocode}
\usepackage{booktabs} 
\usepackage[colorinlistoftodos]{todonotes}
\usepackage{hyperref}
\hypersetup{
    colorlinks=true,
    linkcolor=blue,
    citecolor=blue,
    urlcolor=cyan
    }

\definecolor{cartPoleCol}{HTML}{7fc97f}
\definecolor{acrobotCol}{HTML}{beaed4}
\definecolor{cartCenteringCol}{HTML}{fdc086}
\definecolor{pendulumCol}{HTML}{ffff99}
\definecolor{mountainCarCol}{HTML}{386cb0}
\definecolor{mountainCarContinuousCol}{HTML}{f0027f}
\definecolor{tblGrey}{HTML}{C0C0C0}

\def\makeheadbox{\relax} 
\usepackage{amsfonts} 

\begin{document}

\title{Evolving Hierarchical Memory-Prediction Machines in Multi-Task Reinforcement Learning
}

\titlerunning{Hierarchical Multi-Task Reinforcement Learning}        

\author{Stephen Kelly    \and
        Tatiana Voegerl \and
        Wolfgang Banzhaf \and
        Cedric Gondro 
}

\authorrunning{Kelly, Voegerl, Banzhaf, and Gondro} 

\institute{BEACON Center for the Study of Evolution in Action\\
               Michigan State University\\
              \email{\{kellys27\}@msu.edu}
}

\date{Received: date / Accepted: date}

\maketitle
\newacro{GA}[GA]{Genetic Algorithm}
\newacro{GP}[GP]{Genetic Programming}
\newacro{NPC}[NPC]{Non-Player Character}
\newacro{RL}[RL]{Reinforcement Learning}
\newacro{MTRL}[MTRL]{Multi-Task Reinforcement Learning}
\newacro{LSTM}[LSTM]{Long Short-Term Memory}
\newacro{ERL}[ERL]{Evolutionary Reinforcement Learning}
\newacro{TPG}[TPG]{Tangled Program Graph}
\newacro{VM}[VM]{Virtual Machine}
\newacro{PO}[PO]{Partially-Observable}
\newacro{FO}[FO]{Fully-Observable}
\newacro{GPU}[GPU]{Graphics Processing Unit}

\begin{abstract}
A fundamental aspect of behaviour is the ability to encode salient features of experience in memory and use these memories, in combination with current sensory information, to predict the best action for each situation such that long-term objectives are maximized. The world is highly dynamic, and behavioural agents must generalize across a variety of environments and objectives over time. This scenario can be modeled as a partially-observable multi-task reinforcement learning problem. We use genetic programming to evolve highly-generalized agents capable of operating in six unique environments from the control literature, including OpenAI's entire Classic Control suite. This requires the agent to support discrete and continuous actions simultaneously. No task-identification sensor inputs are provided, thus agents must identify tasks from the dynamics of state variables alone \textit{and} define control policies for each task. We show that emergent hierarchical structure in the evolving programs leads to multi-task agents that succeed by performing a temporal decomposition and encoding of the problem environments in memory. The resulting agents are competitive with task-specific agents in all six environments. Furthermore, the hierarchical structure of programs allows for dynamic run-time complexity, which results in relatively efficient operation.

\end{abstract}

\section{Introduction}\label{sec:intro}

Life is full of new situations and challenges that pose a high degree of uncertainty for 
organisms. In many cases, this uncertainty can only be mitigated through trial-and-error interaction with the environment. For example, the challenge of learning to walk or ride a bike cannot be solved by studying a dataset of examples for how one should map feelings to muscle movements in every possible situation. No such dataset, or model of behaviour, exists. \ac{RL} is a general process through which living organisms and computational machines can manage this type of uncertainty through trial-and-error interaction with the problem environment over time \cite{Neftci2019}, \cite{sutton98}. In machine \ac{RL}, the learning agent is represented by a \ac{VM}, and time is divided into discrete steps. At each timestep, the agent observes its environment through sensor inputs, takes an action that changes the state of the environment, and receives a feedback signal that describes the desirability of its current situation. The goal is to develop agent behaviours that map observations to actions such that the summed feedback, or reward, over all timesteps is maximized, see Figure \ref{fig:rl}. 

\subsection{Multi-Task Reinforcement Learning Environment}\label{sec:mtrl}
The unique \ac{MTRL} environment formulated in this work includes partially-observable versions of the following 6 widely-used \ac{RL} benchmarks from the literature \cite{sutton98}: CartPole, Acrobot, CartCentering, Pendulum, MountainCar, and MountainCarContinuous, Figures \ref{fig:cartPole} to \ref{fig:mountainCar}. These are dynamic control problems with between 2 and 4 state variables and a mix of discrete and continuous action spaces. For example, in the CartPole task (Figure \ref{fig:cartPole}), a pole is attached by an un-actuated joint to a cart, which moves Left or Right along a frictionless track. The state of the system at each timestep, $\vec{s}(t)$, is described for 4 variables including the cart position ($x$), cart velocity ($\dot{x}$), pole angle ($\theta$), and pole velocity at the tip ($\dot{\theta}$). The system is controlled by applying a force of +1 or -1 to the cart. The pole starts nearly upright, and the goal is to prevent it from falling over. A reward of +1 is provided for every timestep that the pole remains upright. The episode ends when the pole is more than 15 degrees from vertical, or the cart moves more than 2.4 units from the center. Complete details about all tasks and implementations used in this work can be found in OpenAI Gym's Classic Control Suite \cite{brockman2016openai}. 

Critical characteristics of these \ac{RL} problems can be summarized as the following:

\begin{figure}[!htb]
	\centering
	\subfigure[CartPole]{\includegraphics[height=3cm]{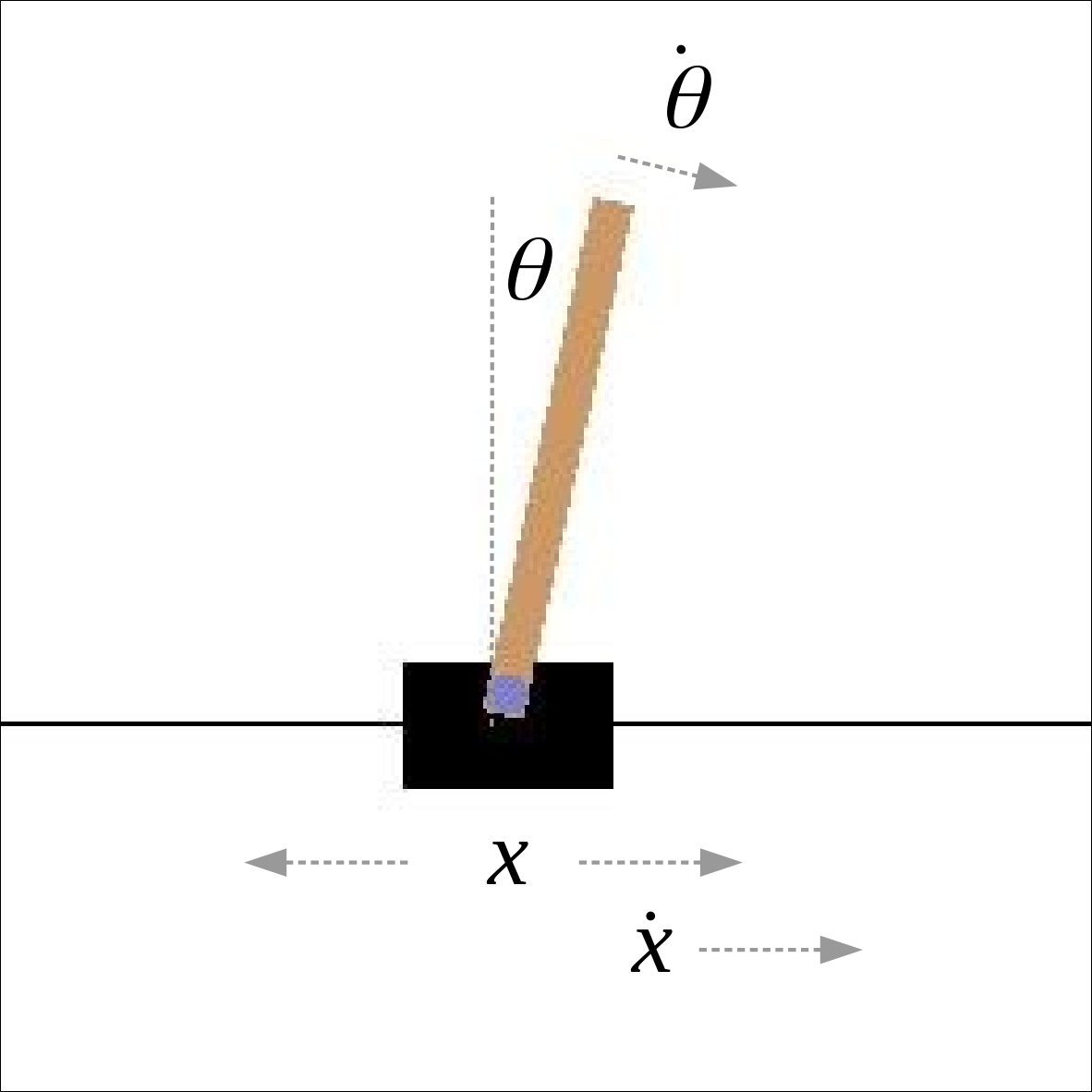}\label{fig:cartPole}}
	\hspace{0.25cm}
	\subfigure[Acrobot]{\includegraphics[height=3cm]{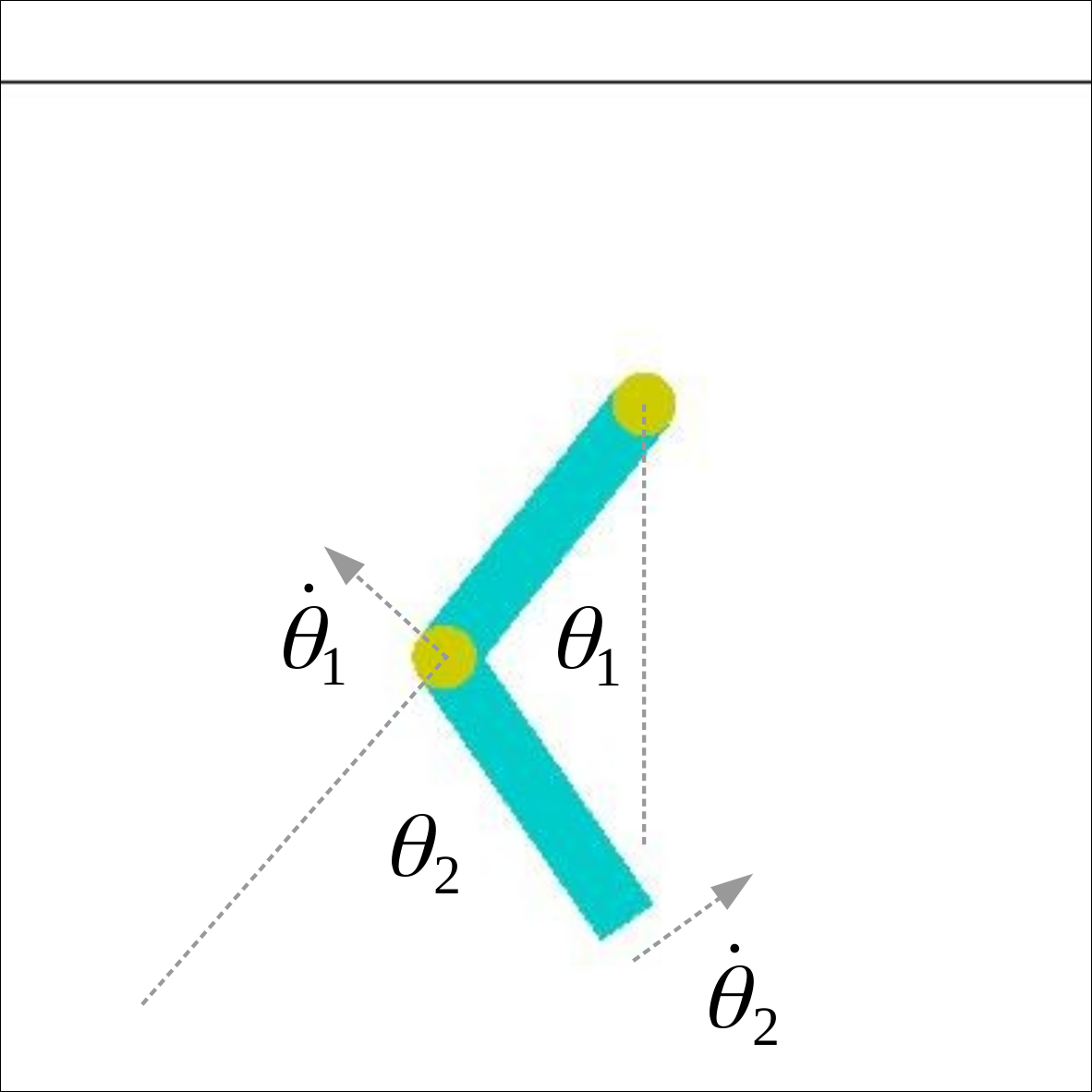}\label{fig:acrobot}}
	\hspace{0.25cm}
	\subfigure[CartCentering]{\includegraphics[height=3cm]{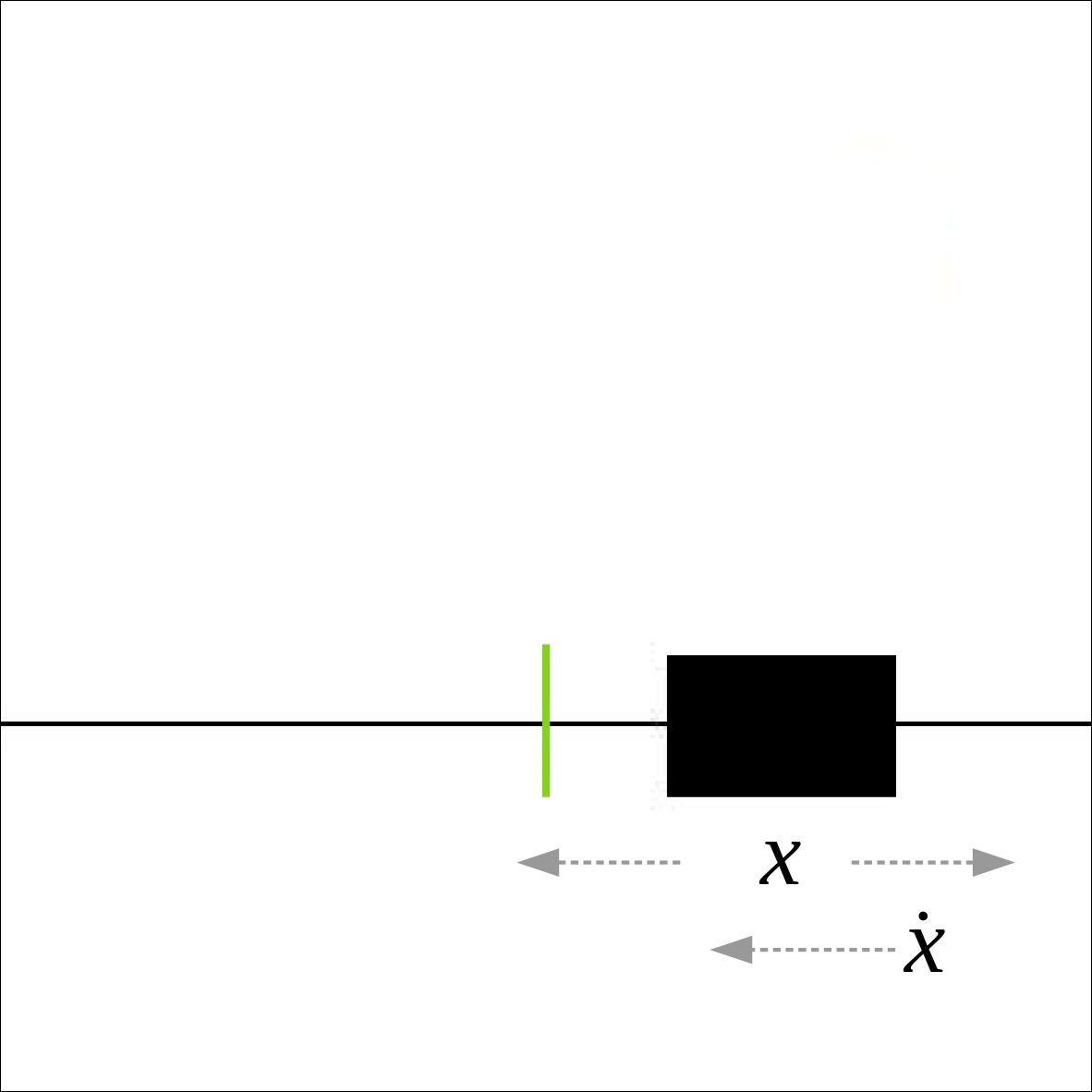}\label{fig:cartCentering}}
	\hspace{0.25cm}
	\subfigure[Pendulum]{\includegraphics[height=3cm]{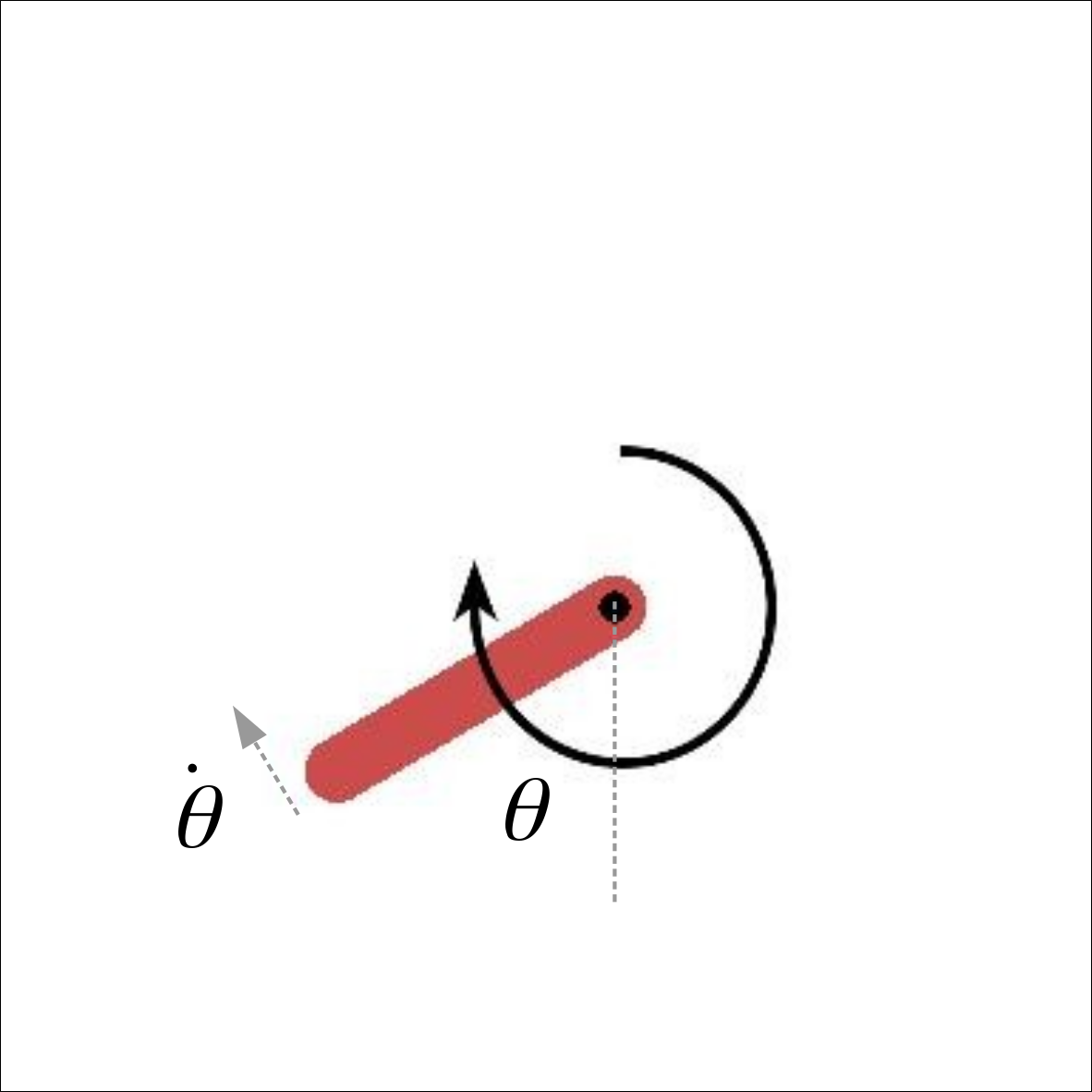}\label{fig:pendulum}}
	\hspace{0.25cm}
	\subfigure[MountainCar]{\includegraphics[height=3cm]{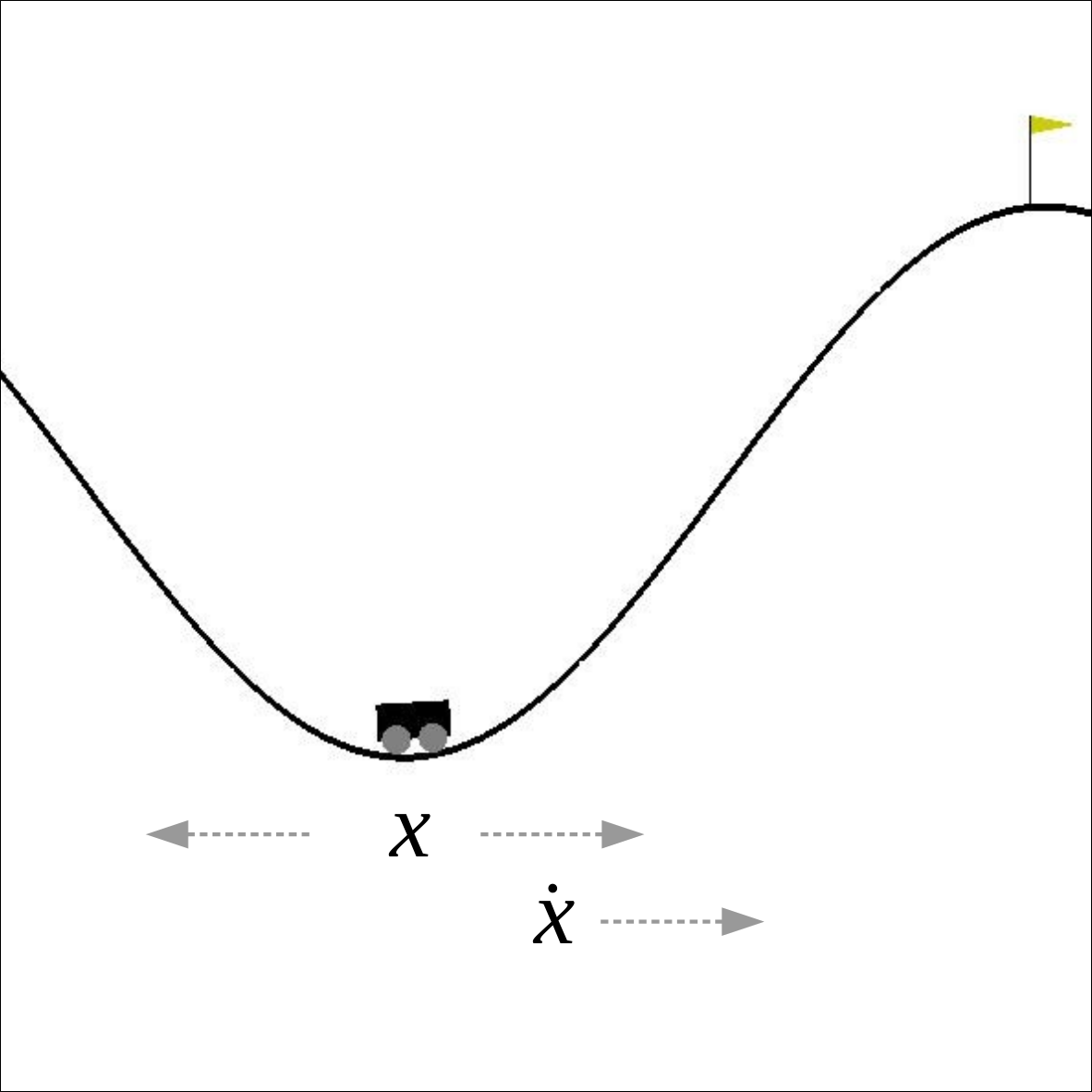}\label{fig:mountainCar}}
	\hspace{0.15cm}
	\subfigure[RL Loop]{\includegraphics[height=2.5cm]{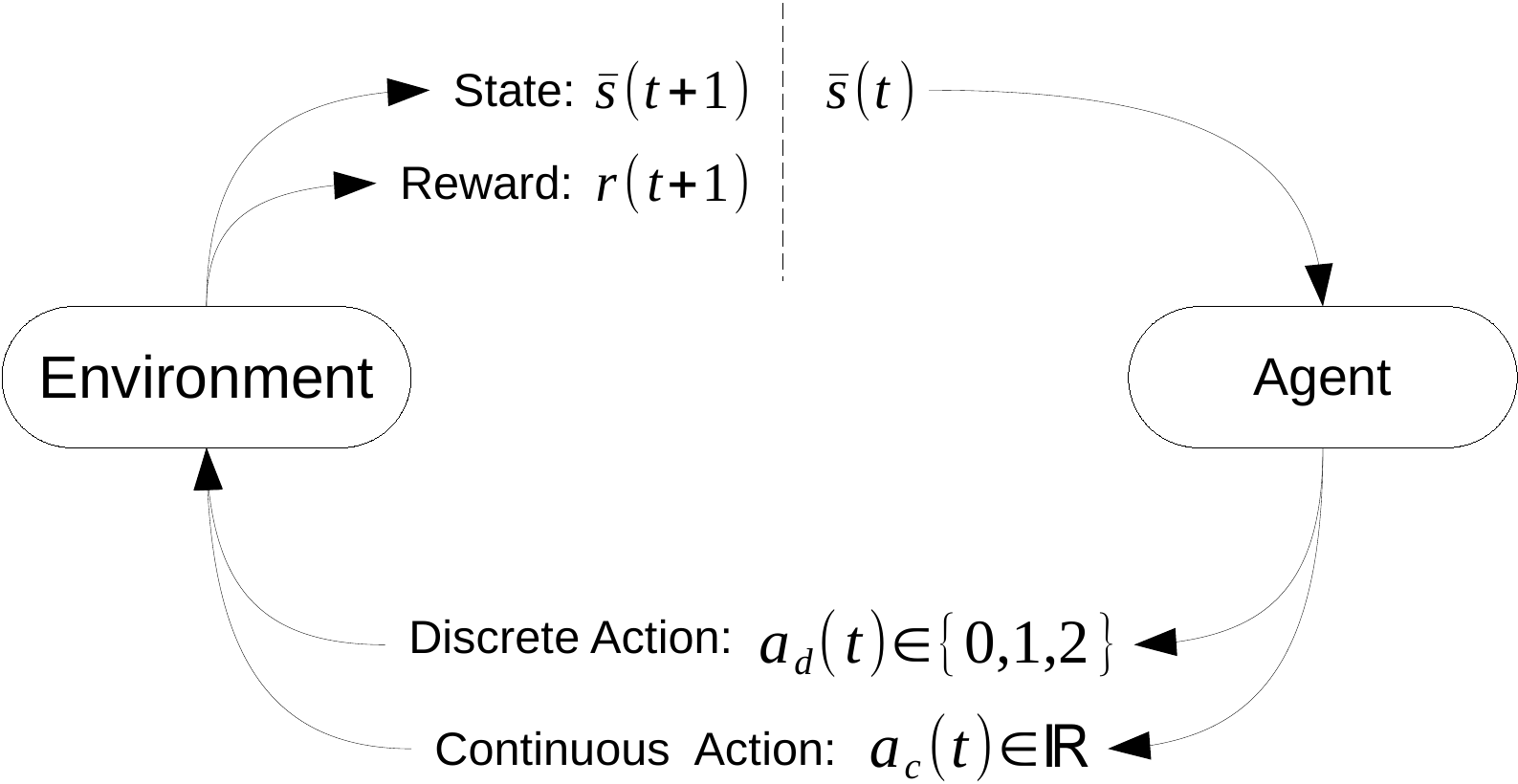}\label{fig:rl}}
	\caption{Classic Control task environments used in this work. For complete details on each task, see \cite{brockman2016openai}}
	\label{fig:tasks}
\end{figure}

\paragraph{Episodic Interactions.} Agent-environment interactions are episodic. Each interaction begins in an initial state of the environment (often a stochastic sampling of the state variables, $\vec{s}$, and continues until a terminal state is reached or a time constraint is exceeded. The quality of an agent's behaviour can be characterized by the sum total of rewards received over the course of an episode. 
    
\paragraph{Delayed Feedback.} The feedback signal is often time delayed, implying that the effect of each action may not be apparent until several timesteps in the future. Furthermore, actions that result in neutral or negative immediate feedback may still contribute to a successful overall outcome. As such, agents that learn \textit{online} by adjusting their behaviour relative to the immediate reward after each action will need to cope with the \textit{temporal credit assignment} problem \cite{Holland:1985}. Conversely, agents that only modify their behaviour once the final episode outcome is known can effectively side-step this issue. For example, decision-level credit in evolutionary \ac{RL} with \ac{GP} is applied implicitly, since agents that make better decisions will receive higher total reward (fitness) and produce more offspring, thus evolution directs the search in favour of agents that make individual decisions that contribute to a positive overall outcome. This approach to \ac{RL} is broadly known as \textit{policy search}. 
    
\paragraph{Mixed Discrete and Continuous Actions.} Depending on the problem, actions may be discrete valued, continuous, or both. For example, in the CartPole task described above, the agent controls the system with a bang-bang force by selecting from 2 discrete actions (1 or -1). In the Pendulum task, the agent must swing a pendulum upright and balance it by supplying a continuous torque value applied to the joint.  Other examples include learning to play Atari video games, where the agent must select from a set of 18 discrete actions corresponding to joystick positions \cite{mnih15}. In the challenging \ac{RL} benchmark of RoboCup soccer, the agent may be required to select which teammate to kick the ball to \textit{and} provide a continuous value describing how hard to kick \cite{fu2019deep}. Continuous action spaces introduce non-trivial design choices for the RL practitioner \cite{annurev-control},\cite{lillicrap2015continuous},\cite{EVCO_a_00080}. For example, continuous control problems cannot be solved by simply discretizing the action space due to the exponentially large number of bins over which policies would have to be learned \cite{metz2017discrete}.

\paragraph{Partial Observability} The agent observes its environment at each timestep $t$ through a sensory interface that provides a set of \textit{state variables} , $\vec{s}(t)$. In many cases, these observations do not contain all the information required to determine the best action, i.e. the envrionment is partially-observable. For example, consider a maze navigation task in which $\vec{s}(t)$ does not contain a global map of the maze, or an environment that contains entities in motion but does not provide their velocity, which is the case for all the control problems considered in this work. In partially-observable environments, the agent is required to identify and store salient features of $\vec{s}$ in memory over time, encoding a representation of the environment that captures temporal properties of the current state \cite{gomez05}. Thus, part of the agent's behaviour must be dedicated to \textit{active} perception\cite{oh2016control}: constructing and managing a representation of the environment in memory. This is distinct from purely reactive agents that define a direct mapping from state to action without any temporal integration of experience.  In RL, agents are also active in the sense that their action choices influence the state of the system and hence their experience of the environment. They must balance exploration vs. exploitation: exploring enough of the environment to build a useful internal model, while also using the model to select actions that optimize their objective.

\paragraph{Non-Stationary, Multi-Task Environments.} The environment defines a transition function that maps the state of the system at time $t$, $\vec{s}(t)$, and the action provided by the agent, $a(t)$, to the next state and reward, $\vec{s}(t+1)$ and $r(t+1)$. The real world is highly dynamic, and realistic machine \ac{RL} can model this by designing non-stationary benchmark environments in which the transition function and/or the reward function changes over time. Video games are a prime example of non-stationary tasks: as the player interacts with the game, new 'levels' of play are encountered and the physics of the simulation change (e.g. entities react differently and move faster) such that gameplay becomes increasingly challenging \cite{yannakakis2018}. Importantly, the agent should be able to adapt to environmental changes without forgetting behaviours that are intermittently important over time. Managing multiple modes of behaviour is the central focus of \ac{MTRL}. More broadly, the goal of \ac{MTRL} is to build generalized agents capable of operating in multiple environments without requiring an oracle to identify which situation is currently being experienced. That is, $\vec{s}(t)$ does not contain information which would explicitly identify the task. At any point in time, the agent must infer which task environment it is interacting with by observing how the state variables change over time, and then behave in a manner that satisfies the objective of the task \cite{electronics9091363,kelly18b}.
    
\begin{table}[!htb]
\scriptsize
\begin{tabular}{lcccc|cccc}
Task            &\multicolumn{4}{c}{State Variables}        &\multicolumn{3}{c}{Disc. Act $a_d \in \{0,1,2\}$}                      &Cont. Act $a_c \in \mathbb{R}$\\
\hline
                &              &          &            &              &                  &  Mapping to Force    &          &  \\ 
                & 0             & 1         & 2            & 3             & 0                 &1      & 2         &  \\
\hline
CartPole        & $x$           & $\theta$  & \cellcolor[HTML]{C0C0C0}$\dot x$      & \cellcolor[HTML]{C0C0C0}$\dot\theta$  & $1$  & $prev$  & $-1$     &    \\
Acrobot         & $\theta_1$    & $\theta_2$& \cellcolor[HTML]{C0C0C0}$\dot\theta_1$& \cellcolor[HTML]{C0C0C0}$\dot\theta_2$&                                   &       &           & $Torque=a_c$ \\
CartCentering   & $x$           & rand & \cellcolor[HTML]{C0C0C0}$\dot x$      &               & $1$                              & $prev$  & $-1$     & \\
Pendulum        & $\theta$      & rand & \cellcolor[HTML]{C0C0C0}$\dot\theta$  &               &                                   &       &           & $Torque=a_c$ \\
MountainCar     & $x$           & rand & \cellcolor[HTML]{C0C0C0}$\dot x$      &               & $1$                              & $0$    & $-1$     & \\
MountainCarC.   &  $x$             & rand & \cellcolor[HTML]{C0C0C0}$\dot x$      &               &                               &     &      & $Force=a_c$
\end{tabular}
\caption{Agent-Environment interface, see Figure \ref{fig:rl}. The observable state at time t, $\vec{s}(t)$, contains state variables 0 and 1. The agent cannot observe variables that describe temporal properties of the system (i.e. velocities in grey). To maintain a common 2-input interface for all tasks, in certain cases the second state variable is replaced by a random number in [0,1]. Disc. Act and Cont. Act  describe how discrete and continuous actions are interpreted by each task. $prev$ indicates the previous action is repeated. Blank cells indicate the action is ignored.}
\label{tbl:interface}
\end{table}

In this \ac{MTRL} study, the goal is to build a single agent that can learn to solve all tasks in Figures \ref{fig:cartPole} to \ref{fig:mountainCar} through direct interaction with the environment. Table \ref{tbl:interface} describes a common agent-environment interface used for all tasks. Notice that the state of each system is described by the position and velocity of different entities (Table \ref{tbl:interface}). In this work, the agent is blind to velocity variables, implying that all tasks are partially-observable. In order to solve these problems, agents will need to predict the system velocities by integrating the observable variables over time.  The state observation, $\vec{s}(t)$, contains 2 state variables. Note that neither variable explicitly identifies the task. The observable state variables are normalized to the range $[-1,1]$ to insure that their magnitude cannot be used to identify the task. The agent will need to infer which task it is currently interacting with by observing how the state variables change over time. Finally, the agent must produce 1 discrete action and 1 continuous action at each timestep. CartPole, CartCentering, and MountainCar will respond to the discrete action, while the remaining tasks will respond to the continuous action. This \ac{MTRL} challenge is exceptionally difficult. However, the individual tasks are well-known, tractable \ac{RL} benchmarks. Thus, with this methodology we establish the minimum essential properties for a new \ac{MTRL} testbed. Algorithms evaluated in this tesbed will need to address the following primary challenges of \ac{MTRL}\cite{electronics9091363}:
\begin{enumerate}
    \item \textbf{Scalability} Jointly learning $N$ tasks should not take $N$ times as long as learning each task individually, and the resulting multi-task agent should not be $N$ times as complex.
    \item \textbf{Distraction Dilemma} The magnitude of each task's reward signal may be different, causing certain tasks to appear more salient than others.
    \item \textbf{Catastrophic Forgetting} When learning multiple tasks in sequence, the agent must have a mechanism to avoid unlearning task-specific behaviours that are intermittently important over time.
    \item \textbf{Negative Transfer} If the environments and objectives are similar, then simultaneously learning multiple tasks might improve the learning/search process through positive inter-task transfer. Conversely, jointly learning multiple dissimilar tasks is likely to make \ac{MTRL} more difficult than approaching each task individually.
\end{enumerate}

\subsection{Tangled Program Graphs and Emergent Modularity}\label{sec:tpg-mod}
\ac{TPG} is a Genetic Programming (GP) framework  which incrementally builds computational organisms from multiple subsystems which were initially developed independently, akin to compositional evolution \cite{watson_modular_2005}. In doing so, TPG automates two critical properties of such a system: 1) The identification of stable building blocks, or subsystems; and 2) Establishing the nature of the interaction among subsystems within a hierarchical organism, or \textit{module interdependence}. 

Relative to the first property to be automated, i.e. discovery of stable building blocks, Herbert Simon \cite{Simon62} suggested that the presence of stable intermediate structures speeds up evolution by providing building blocks from which increasingly complex hierarchies may be constructed. Put simply, Simon points out that if a complex system is built from structurally modular building blocks, its development is less likely to require a restart from scratch should an error be introduced during construction (see Simon's famous Watchmaker's Parable for an illustrative example of this concept). In other words, modularity helps promote stability in an evolving organism, preventing a particular genome from being a "House of Cards" \cite{kingman_1978} in which a single variation might bring it tumbling down. Ultimately, Simon's suggestion is that modular systems are more \textit{evolvable}, that is, more capable of continuously discovering new organisms with higher fitness than their parents. This theory has been investigated widely among evolutionary biologists \cite{Nedelcu02,Wagner96,Yang01}.

As for the second property to be automated, module interdependence, Watson et al. \cite{watson_modular_2005} emphasize that structural modularity (i.e. structural complexity encapsulated such that dependencies between subsystems are weaker than dependencies within subsystems) does not imply independence of subsystems. Specifically, functional interdependence among subsystems is critical for hierarchies in which all levels of organization are meaningful. Simply accumulating multiple building blocks into an aggregate system does not capture the full potential of modularity. Module interdependence is essential for emergence because without meaningful interdependence, a hierarchy of subsystems is nothing more than the sum of its parts. Watson argues that systems with strong module interdependence are evolvable under certain conditions, namely compositional evolution.

TPG has leveraged emergent modularity in hierarchical model building to make a variety of contributions in the context of visual Reinforcement Learning (RL). In the Atari video game testbed, TPG evolved game-playing agents that match the quality of solutions from a variety of deep learning methods \cite{kelly18}. More importantly, TPG agents were less computationally demanding and required fewer calculations per decision than any of the other methods. This efficiency is possible because 1) the hierarchical complexity of each organism is a property that emerges through interaction with the problem environment, rather than being fixed a priori, as was the case for deep learning, e.g. \cite{mnih15}; and 2) subsystems within a TPG organism typically specialize on different parts of the visual input space, thus only subsets of the overall organism require execution at any given point in time.

Modularity and specialization also allow TPG to support transfer learning in challenging RL problems \cite{kelly18b}. In this case, solutions initially evolved for simple subtasks can be reused within hierarchical organisms in order to improve learning in a more complex task. The resulting agents achieve state-of-the-art levels of play in RoboCup Half-Field Offense and surpass scores previously reported in the Ms. Pac-Man literature while employing less domain knowledge during training. Again, the highly modular organisms are shown to be significantly more efficient than state-of-the-art solutions in both domains. 

Finally, modularity and specialization are also useful in dynamic environments where the distribution in sensory inputs may change drastically over time. When forced to switch randomly between multiple Atari game titles throughout evolution, TPG can evolve solutions to multiple titles simultaneously with no additional computational cost \cite{kelly18}. In this case, modularity was critical to avoid unlearning  or "catastrophic forgetting" \cite{Kirkpatrick3521} of behaviours that are intermittently important over time.

\subsection{Modular Memory Models}
All the work outlined in Section \ref{sec:tpg-mod} was conducted using an early version of TPG in which organisms were \textit{stateless}. That is, even though agents operated in episodic, sequential decision-making environments involving hundreds or thousands of timesteps, the agents were purely reactive. They had no temporal memory mechanism to enable the integration of experience over time. This is a serious limitation in partially-observable tasks in which it is impossible to retrieve complete information about the state of the environment from a single observation. More recently, multiple models have been proposed which support temporal memory sharing among subsystems within TPG organisms, allowing agents to operate in sequential decision-making environments with partial observability at multiple time scales \cite{kelly20a,smith19b,smith19a}. Examples from the deep learning community have also demonstrated that modularity and specialization lead to improved generalization in dynamic tasks that require temporal reasoning \cite{banino2020memo,goyal2019recurrent}.

\section{Research Objectives}
Section \ref{sec:tpg-mod} established the capabilities of TPG for evolving hierarchical/modu-lar agents in high-dimensional (e.g. visual) RL environments with discrete action spaces. The approach has recently been extended to incorporate temporal memory mechanisms that enable operation in environments with partial-observability at multiple time scales. The work herein is an extension of our study published at GECCO 2020 \cite{kelly20b}. The first objective of this initial study was to propose a highly-modular memory structure that manages temporal properties of a task and enables operation in problems with continuous action spaces. This significantly broadens the scope of real-world applications for TPGs, from symbolic regression to time series forecasting.

TPG's success in high-dimensional RL was due in part to its capacity to adaptively decompose the input space such that individual subsystems within an organism could specialize their role relative to small subsets of the input space, or \textit{spatial} decomposition \cite{kelly18}. The second objective of our initial study was to examine how the modular memory mechanism allows organisms to achieve a \textit{temporal} problem decomposition. This is significant because temporal problem decomposition is likely beneficial in dynamic, non-stationary environments. Examples of this include \ac{MTRL}, as well as time series forecasting or streaming data classification tasks when the underlying process generating the data stream changes significantly over time \cite{Agapitos2012,heywood_evolutionary_2015}.

Putting these developments together, the overall purview of this work is to demonstrate how \ac{TPG} can be used to build hierarchical memory-prediction machines that address the \ac{MTRL} challenges outlined in Section \ref{sec:mtrl}. First, we test the hypothesis that TPG's shared memory framework \cite{kelly20a,kelly20b} can be further extended to support continuous and discrete action spaces \textit{and} temporal memory management simultaneously. Next, we propose that a fundamental property of a successful multi-task behavior is its ability to hierarchically decompose the problem.  To this end, we show that \ac{TPG} can evolve hierarchical multi-task behaviors by combining several agents which were initially adapted independently. Over time, a collective behavior emerges that builds on the individual specializations of multiple agents. Finally, we evaluate TPG's ability to manage partial-observability in multi-task environments. To achieve this, we examine how TPG's modular memory mechanism \cite{kelly20a,kelly20b} allows agents within a hierarchical \ac{VM} to share temporal information and collectively build a shared representation of environmental state. Critically, both hierarchical problem decomposition and shared memory management are emergent properties of an open-ended evolutionary system.

The remainder of this paper is organized as follows: Section \ref{sec:related} reviews recent work in \ac{MTRL}. Section \ref{sec:tpg} provides a detailed description of the extended TPG algorithm. An empirical evaluation is provided in Section \ref{sec:eval}. We evaluate TPG in the context of learning 6 unique environments from the control literature, including OpenAI's entire Classic Control suite. This requires the agent to support discrete and continuous actions simultaneously. No task-identification inputs are provided, thus agents must identify tasks from the dynamics of state variables alone \textit{and} define control policies for each task. We show that emergent hierarchical structure in the evolving programs leads to multi-task agents that succeed by performing a temporal decomposition/encoding of the problem environments in memory. The resulting agents are competitive with task-specific agents in all 6 environments. Furthermore, the hierarchical structure of programs allows for dynamic run-time complexity, which results in relatively efficient operation. Section \ref{sec:conclusion} concludes the paper and provides an outlook to future work.

\section{Related Work in Deep Learning}\label{sec:related}
Two broad research questions are explored in the \ac{MTRL} literature: 1) How to support knowledge sharing \textit{across} multiple related tasks; and 2) How to support multiple unrelated or competing tasks by decomposing the overall problem and problem solver (agent). 

\subsection{Shared Representations and Manual Decomposition}
In deep learning, support for shared knowledge primarily takes the form of learning shared feature representations. That is, how can networks be developed such that weight parameters are general enough to model features relevant to multiple tasks. D'Eramo et. al \cite{DEramo2020Sharing} recently formulated proofs that this approach can lead to gains in performance and sample efficiency when compared to single-task learning. However, only part of the network was shared among tasks. The multi-task problem is manually decomposed in order to design a network with task-specific input and output layers for each task. Knowledge of which task the network is currently interacting with is required to select which task-specific network components to activate at any timestep. Furthermore, a separate replay memory is required for each task, incurring a significant memory overhead compared to single-task learning. 

Policy distillation \cite{rusu2016policy} is another deep learning approach to developing shared representations for \ac{MTRL}. In this case, multiple pre-trained, single-task DQN agents \cite{mnih15}, called \textit{teachers}, are used to generate a multi-task replay memory (i.e. a dataset) of example $<state,action>$ pairs. A \textit{student} network is then trained from the replay memory using supervised learning. The student can effectively model the behaviour of multiple DQN agents. Furthermore, the student is typically a simpler network, thus policy distillation can result in a scaled-down, faster \ac{MTRL} agent with performance comparable to multiple DQN teachers. However, pretraining a single-task DQN teacher for each task incurs a significant computational cost. Furthermore, multi-task decomposition is pre-configured manually: the student network included a separate output layer trained for each task, once again implying that a task label is required during model deployment to select the correct output layer at each timestep.

IMPALA and PopArt \cite{Hessel2019} are deep learning methods that leverage a distributed actor-learner architecture to propose a \textit{scalable} method of learning shared representations in \ac{MTRL}. In short, a centralized \textit{learner} network acts as a shared parameter server from which multiple \textit{actor} networks can copy parameters before going off to interact with multiple unique task environments in parallel. Each actor's experience ($<state, action>$ pairs) is periodically (asynchronously) integrated back into the learner's shared representation. PopArt included a method of normalizing the rewards over the entire task set, thus improving over IMPALA by avoiding the distraction dilemma. The entire network architecture is shared among all tasks, implying that the power of these methods lies in their ability to learn generalized feature representations that captured salient properties of \textit{all} tasks. That is, there is an underlying assumption that all of the tasks have something in common, and therefore problem decomposition is not given significant attention. However, no task label is required to switch between task-specific modules. The network input consisted solely of the $96 \times 72$ pixel matrix (i.e. the game screen), implying that the network could infer the task without a access to a label. Finally, the network architecture included a \ac{LSTM} \cite{Greff2017} module. As such, the method could be applied to partially-observable environments such as the first-person 3D DeepMind Lab benchmark suite \cite{beattie2016}. However, no ablation study was performed to confirm the significance of the \ac{LSTM}.

\subsection{Shared Representations and Automatic Decomposition}

Methods that attempt \textit{automatic} problem decomposition typically incorporate some form of modularity to build prediction machines with diverse structural components that can specialize on subsets of the overall problem. Soft Modularization \cite{yang2020multitask} is one such approach.  In this case, a \textit{base policy} network, which maps $\vec{s}(t)$ to an action, is trained together with a \textit{routing} network. At each timestep, the routing network is given a 1-hot task embedding (i.e. task label) and probabilistically selects a route through the the base policy network. In effect, the routing network dynamically selects which \textit{modules} in the base policy network should be active for the task at hand. The architecture for both networks is predefined, thus the nature of the modularity is not emergent. However, the base policy design provides a modular template such that the routing network can effectively learn how to decompose the multi-task problem within specialized structural modules which are dynamically switched in and out of the execution path at run-time. This improves positive inter-task transfer compared to networks with fixed routing because modules that specialize at specific aspects of the problem can be switched in when they are required and switched out when their (over) specialization might result in negative transfer. Dynamic routing also improves efficiency because only part of the overall network is executed at each timestep.  The primary limitation of Soft Modularization as is that knowledge of the active task label is required as input to the routing network.    

Progressive Neural Networks \cite{rusu2016} take hand-designed modularity to an extreme, dedicating an entire network to each task. The framework is designed for multi-task learning scenarios in which a sequence of tasks is pre-defined and the machine learns each new task in sequence. A new network is added for each task and the weights of all previous networks (columns) are frozen to avoid catastrophic forgetting. Lateral connections connect each frozen network to all subsequent nets. The final machine solves up to 4 Atari tasks, and it is shown that positive transfer from previous networks/tasks can significantly accelerate learning new tasks. The primary limitation of Progressive Neural Networks is scalability because a new network is added for each new task. Furthermore, while all networks process $\vec{s}(t)$ at each timestep, the output of only one net must be selected using knowledge of the active task label. Elastic Weight Consolidation (EWC) \cite{Kirkpatrick3521} improved on this by using a single network for continual learning of multiple tasks. The algorithm slows down updates on certain weights based on how important they are to previously seen tasks. Furthermore, a task-recognition model was incorporated to infer which task is being performed and automatically manage which sets of weights to protect at any given time. A DQN agent augmented with EWC was able to learn up to 10 Atari games. However, it did not reach the score that would have been obtained by training ten separate DQNs. Furthermore, DQN side-steps the issue of partial-observability by using an autoregressive state representation. In short, \textit{frame stacking} is employed such that $\vec{s}(t)$ contains a hard-coded historical window of the 4 most recent state observations (See \cite{mnih15}). As such, no temporal memory mechanism is required to infer short-term temporal properties of the environment such as the directional velocity of moving game entities. This approach to dealing with partial-observability is limited because a designing a temporal sliding window, or autoregressive state, relies on the experimenter's intuition/assumptions about the environment, and can only mitigate partial-observability within the fixed window. Furthermore, the machine is unable to adapt this window if the properties of the task change over time. 

PathNet \cite{fernando2017} is an approach to sequential multi-task learning which evolves subnetworks within a super network, essentially discovering how to reuse parameters from previous tasks while learning new ones. Learning takes place over two distinct timescales: Online gradient descent adjusts the weights of "active" subnets as they interact with the environment. A \ac{GA} is used to discover which parts from a template super network to use within each subnet. As new tasks are introduced, the best subnets and their weight parameters from the previous task are frozen. This mechanism supports multi-task parameter reuse without catastrophic forgetting, and demonstrated positive inter-task transfer.  However, PathNet was only evaluated on sequential learning of 2 Atari and Labyrinth games. Furthermore, the network architecture still included a separate output layer for each task. As such, the networks have no mechanism to identify which environment they are interacting with, and a task label is again required. 

In summary, there has recently been a surge of work in \ac{MTRL}, but to date there has not been significant progress made on approaches that address all the fundamental properties that make \ac{MTRL} challenging. The motivation for this study is to fill this gap with an evolutionary approach to \ac{MTRL} in which: 1) Agent complexity scales through interaction with the environment, and the \textit{run-time} complexity of the agents does not grow linearly with the number of tasks; 2) The agent's multi-task behaviour includes task-recognition capability, removing the need for an oracle to provide the current task label; 3) The environments are partially-observable and require agents to support temporal memory. 

\section{Algorithm Description}\label{sec:tpg} 

The algorithm investigated in this work is an extension of Tangled Program Graphs \cite{kelly18}. TPG was initially designed for RL tasks in which solutions map sensor inputs to a set of discrete actions. This work represents the first time the method has been used to build programs capable of operating in discrete-action and continuous-action RL environments simultaneously, which is achieved through an extension of the shared memory mechanism introduced in \cite{kelly20a}. This section outlines the extended algorithm, paying specific attention to two critical components: 1) How memory is shared among individual programs in a team-based model; and 2) How multiple independent teams are adaptively combined into a hierarchical organism, or \textit{program graph}, through compositional evolution. All source code is publicly available \cite{tpgSource}.

\subsection{Coevolving Independent Teams}
\begin{figure}[!htb]
	\centering
	\includegraphics[width=.7\textwidth]{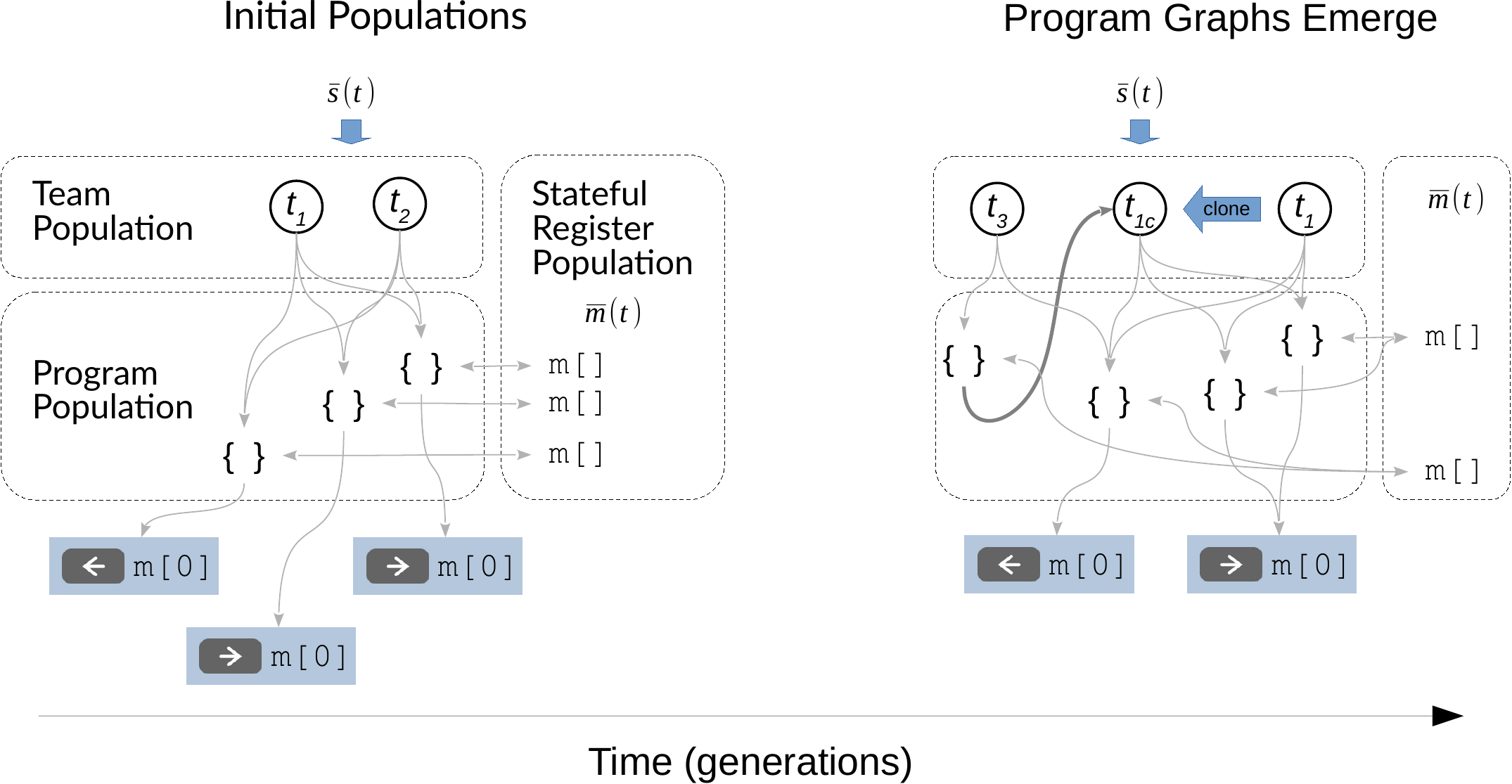}
	\caption{Illustration of the relationship between teams, programs, and shared memory in TPG. Initially, all programs are leaf nodes. Over time, program action pointers may be modified to refer to other teams and \textit{program graphs} emerge. When a team is subsumed into a program graph, it is cloned and the clone ($t_{2c})$) becomes an internal node. See Section \ref{sec:compositional} for details.}
	\label{fig:tpg}
\end{figure}

A \textit{team of programs} is the basic representation for a stand-alone agent in TPG. Each team defines a group of programs that \textit{collectively} map input state at time $t$, $\vec{s}(t)$ to a pair of discrete and continuous actions, $<a_{d}, a_{c}>$. Teams can be thought of as vertices in a computational graph where the \textit{edges} are programs that process $\vec{s}(t)$ and produce output, Figure \ref{fig:tpg}. In this work, all programs are linear register machines \cite{brameier07}, see Algorithm \ref{alg:program}. For the purposes of this study, it is important to note that programs contain internal register memory that is \textit{stateless}, that is, reset prior to each execution. Programs also have a pointer to one shared \textit{stateful} memory bank that is only reset at the start of each episode of interaction with the environment. In the case of sequential decision-making tasks where programs are executed multiple times per episode, shared stateful memory allows programs to communicate with each other and to integrate information across multiple timesteps. This is a crucial aspect of behaviour which allows teams to construct an internal world model of partially-observable environments. In this case, the team-based agent must encode salient information from $\vec{s}(t)$ into stateful memory such that it can be reused, in combination with $\vec{s}(t+n)$, when selecting an action a time $t+n$. This is one example of an agent taking an \textit{active} role in its perception of the environment. As we will demonstrate, programs construct their world model dynamically at run-time from the content of temporal memory, $\vec{m}(t)$ \textit{and} the current sensor input, or state $\vec{s}(t)$.

Programs have a dual-purpose role within a team: 
\begin{enumerate}
\item \textbf{Memory Management} In order to manage the content of stateful memory, programs can read from current environmental state, $\vec{s}(t)$, and/or stateful memory, $\vec{m}(t)$, and write to $\vec{m}(t)$; 
\item \textbf{Program Graph Traversal} In the context of a team, programs can be characterized as directed graph edges that dynamically set their weight as a function of $\vec{s}(t)$ and $\vec{m}(t)$. Each team maintains at least two programs, and each program has a pointer to one discrete action (See \ref{fig:tpg}). The team maps $<\vec{s}(t),\vec{m}(t)>$  to a pair of actions $<a_d, a_c>$, by executing all programs in order and then following the path with the largest weight. If the program is a leaf, then $a_{d}$ is the discrete action associated with the winning program, and $a_{c}$ is the content of its shared stateful memory register $m[0]$, i.e., a continuous value left over after all programs have executed (See Algorithm \ref{alg:program}).
\end{enumerate}

\alglanguage{pseudocode}
\begin{algorithm}[!htb]
\small
\caption{Example linear register machine. Each program contains one \textit{private stateless} register bank, $r$, and a pointer to one \textit{shareable stateful} register bank, $m$. $r$ is useful for storing the result of intermediate calculations during execution, and is reset prior to each execution. $m$ is useful for storing values over multiple timesteps and sharing information with other programs. $m$ is only reset (by an external process) at the start of each episode. Note that $\vec{s}(t)$ is read-only, thus the target of each instruction can only be an index into $r$ or $m$. The first operand is alway an index into either $r$ or $m$, while the second operand could reference $r$, $m$, or $\vec{s}(t)$. Programs have two return values (line \ref{ln:rtn}). In this case line \ref{ln:intron} has no effect on the final value of r[0] or any effect on $m$. Ineffective instructions are useful for the evolutionary search, but for efficiency they can easily be identified and skipped during program execution \cite{brameier07}. A complete list of operations and instruction formats appears in Table \ref{tbl:ops}.}
\label{alg:program}
\begin{algorithmic}[1]
    \State $r \leftarrow 0$ \Comment{reset private memory bank}
    \State $m[0] \leftarrow m[7] \div \vec{s}(t)[3]$
    \State $r[1] \leftarrow m[0] \div r[7]$\label{ln:intron}
    \State $r[0] \leftarrow \cos{m[0]}$
    \If {$r[0] < m[2]$} 
    \State $r[0] \leftarrow -r[0]$
    \EndIf
    \State \Return ($r[0]$, $m[0]$) \Comment{(weight, $a_{c}$)}\label{ln:rtn}
\end{algorithmic}
\end{algorithm}

Note that programs simultaneously manage stateful memory and define the appropriate \textit{context} (relative to $\vec{s}(t)$ and $\vec{m}(t)$) in which their action pair should define the agent's output. 
\begin{table}[!htb]
\centering
\footnotesize
  \caption{Operations and instruction formats. Programs encode 16 operations in a 4-bit op-code. In addition, programs have access to 18 constants: \{$-0.9, -0.8, ..., -0.1, 0.1, 0.2, ..., 0.9$\}, included as read-only registers at the end of their private register bank $r$ (See Algorithm \ref{alg:program}). Let $r_x$ and $r_y$ be generic registers.}
  \label{tbl:ops}
  \begin{tabular}{ll}
    \toprule
    Instruction   & Operations    \tabularnewline
    \midrule
    $r_x[i] \leftarrow r_x[i] \circ r_y[j]$             & $\circ \in \{+,-,\times,\div,x^{y}\}$ \tabularnewline
    \midrule
    \multirow{2}{*}{$r_x[i] \leftarrow \circ(r_y[j])$}  & $\circ \in \{\cos,\ln,\exp,\sqrt,\sin\}$ \tabularnewline
                                                        & $\circ \in \{\tanh,x^{2}, \left|x\right|,x^{3} \}$ \tabularnewline
    \midrule
    IF $(r_x[i] \circ r_y[j])$ THEN $r_x[i] \leftarrow -r_x[i]$ & $\circ \in \{<,>\}$ \tabularnewline
  \bottomrule
\end{tabular}
\end{table}

Teams, programs, and shared memory registers are each stored in separate populations and coevolved. Evolution is driven by a generational GA in the following sequence of steps (parameters listed in Table \ref{tbl:param}):

\begin{enumerate}
    \item \label{ga-init} \textbf{Initialization} Evolution begins with a population of $R_{size}$ stochastically generated teams. Each team contains at least $tmSize_{init}$ new programs which are initialized with a unique memory bank (i.e. each initial team has a unique complement of $tmSize_{init}$ programs, and each program has a unique memory pointer), Figure \ref{fig:tpg}. Programs are initially all leaf nodes.
    
    \item \label{ga-gen} \textbf{Generate Offspring} Let $\mathbb{P}$ be the power set of all task combinations. For 6 tasks, $\mathbb{P}$ will contain 63 unique task sets. For each set $s \in \mathbb{P}$, the process for generating team offspring will create $n_{elite}$ new root teams. To create each new root, the process uniformly samples two teams, $parent_1$ (always a root team) and $parent_2$. Crossover is applied with probability $p_x$. When no crossover is applied, $parent_1$ is cloned to create a new child team. Otherwise the crossover operator begins by creating an empty child team. Shared memory implies that the order of program execution within a team potentially impacts the outcome. To avoid disrupting this order, the crossover operator interleaves programs from $parent_1$ and $parent_2$ \textit{in order} within the child, where each parent program is copied to the child with $50\%$ probability, Figure \ref{fig:team-cross}. Mutation operators are then applied to the child team, as listed in Table \ref{tbl:param}. Team mutation operators may modify the discrete action and memory pointers, modify the program order, and add, remove, and modify programs in the team. In short, team complement, program length and content, and the degree of memory sharing are all adapted properties. Further details on TPG's variation operators are available in \cite{skellyphd}. 
    
    \begin{figure}[!htb]
	\centering
	\includegraphics[width=.4\textwidth]{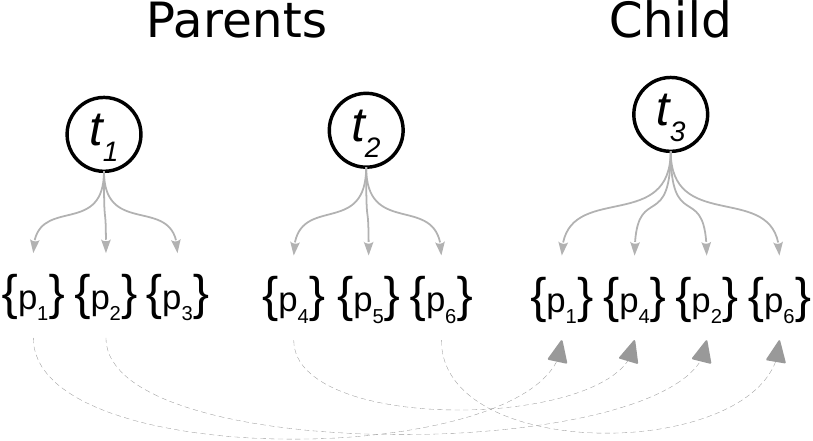}
	\caption{Illustration of team crossover operator. Each parent program is copied to the child with $50\%$ probability. Parent programs are interleaved within the child, maintaining their original ordering. }
\label{fig:team-cross}
\end{figure}
    
    \item \label{ga-eval} \textbf{Evaluation} Every root team in the population represents a stand-alone agent. Thus, every \textit{new} root team (created in the previous step) is evaluated in 5 episodes in each task environment. 
    
    \item \label{ga-sel} \textbf{Selection} For each set $s \in \mathbb{P}$ (the power set of task combinations), $n_{elite}$ teams with the highest fitness are designated as survivors and protected from deletion in this generation. For single task sets, team fitness is simply the average reward over 5 episodes in that task. For multi-task sets, team fitness captures how well a team performs on multiple tasks by ranking teams by their weakest task performance in the set. To achieve this, every root team's mean reward on each task is normalized relative to the rest of the current root population. Normalized score for team $tm_i$ on task $t_j$, or $sc^{nrm}(tm_i, t_j)$, is calculated as $(sc(tm_i,t_j) - sc_{min}(t_j))/(sc_{max}(t_j) - sc_{min}(t_j))$, where $sc(tm_i,t_j)$ is the mean score for team $tm_i$ on task $t_j$ and $sc_{min,max}(t_j)$ are the population-wide min and max mean scores for task $t_j$. Multi-task fitness for team $tm_i$ is then $min(sc^{nrm}(tm_i,t_{\{1..n\}})$, or the minimum normalized score for team $tm_i$ over all tasks. $n$ denotes the number of tasks. Thus, multi-task survivors are the teams with the highest minimum normalized fitness over all tasks in each task set. Any team not identified as a survivor in this process is deleted. Note that normalizing rewards is a critical part of quantifying multi-task fitness \textit{and} mitigates the distraction dilemma (See Section \ref{sec:mtrl}). Finally, programs have no individual concept of fitness. After team deletion, programs that are not part of any team are also deleted. As such, selection is driven by a symbiotic relationship between programs and teams: teams will survive as long as they define a complementary group of programs, while individual programs will survive as long as they collaborate successfully within a team. 
    
    \item Got to step 2.
    
\end{enumerate}

\begin{table}[!htb]
\caption{Parameterization of Team and Program populations. $R_{size}$ is the initial number of root teams. $n_{elite}$ is the proportion of root teams to maintain for each task set (See Section \ref{sec:tpg} text). For the team population, $p_x$ is the probability of crossover and $p_{mx}$ denotes a mutation operator in which: $x \in \{d, a\}$ are the probality of deleting or adding a program respectively; $x \in \{m, n, s\}$ are the probality of creating a new program, changing a path-program's action pointer (leaf or team), and changing a program's shared memory pointer respectively. $\omega$ is the max initial team size. For the program population, $p_{x}$ denotes a mutation operator in which $x \in \{delete,add,mutate,swap\}$ are the probality for deleting, adding, mutating, or reordering instructions within a program. $p_{atomic}$ is the probability that a modified action-pointer for a path-program will be atomic (leaf).}
\label{tbl:param}
\centering
\begin{tabular}{llll}
\toprule
\multicolumn{4}{c}{Team population} \tabularnewline 
Parameter           & Value     & Parameter         & Value             \tabularnewline
\hline
$R_{size}$          & 1000      & $n_{elite}$       & 50                \tabularnewline
$tmSize_{init}$     & 10        & $tmSize_{max}$    & $\infty$          \tabularnewline
$n_{elite}$         & 50        & $p_x$             & 0.2               \tabularnewline
$p_{md}$            & 0.7       & $p_{ma}$          &  0.6              \tabularnewline
$p_{mm}$            & 0.2       & $p_{mn}, p_{ms}$  &   0.1             \tabularnewline \hline
\toprule
\multicolumn{4}{c}{Program population} \tabularnewline 
Parameter           & Value     &   Parameter               &  Value    \tabularnewline
\hline
\textit{Size of $r$}& 8         & \textit{Size of $m$}      &  8        \tabularnewline
$progSize_{init}$   & 10        & $progSize_{max}$          &  $\infty$ \tabularnewline
$p_{delete}$        &   0.5     & $p_{add}$                 & 0.4       \tabularnewline
$p_{mutate}$        &   1.0     & $p_{swap}$                & 0.2       \tabularnewline
$p_{atomic}$        &   0.95    &                           &           \tabularnewline
\hline
\end{tabular}
\end{table}

\subsection{Evolving Team Hierarchies}\label{sec:compositional}
When a program is modified by variation operators in Step \ref{ga-gen}, it will remain a leaf with probability $p_{atomic}$, and will otherwise connect to one team from the set of teams present from any previous generation, chosen with uniform probability. These connection mutations are the mechanism by which TPG supports compositional evolution, adaptively recombining multiple (previously independent) teams into variably deep/wide directed graph structures, or \textit {program graphs}, Figure \ref{fig:tpg}. 

Execution of a program graph begins at the root team ($t_3$ in Figure \ref{fig:tpg}), where all programs in the team will execute in order. Graph traversal then follows the program with the largest weight, repeating the execution process at every team along the path until a leaf node is reached. Thus, the program graph computes one path from root to leaf at each timestep, where only a subset of programs in the graph (those in teams along the path) require execution. Note that cycles may appear in the graph structure but are ignored during execution. That is, no team is visited more than once per traversal. If the edge with the largest weight leads to a team that has already been visited, the edge is simply ignored and the program/edge with the next highest weight is considered. Team variation operators are constrained such that each team maintains at least one program that is a leaf node, ensuring an output can always be found.

As hierarchical structures emerge, only root teams (i.e. teams with indegree of 0) define independent agents, and only these root teams are subject to deletion, cloning, and variation. Non-root teams are protected from deletion as long as they are a component of a graph that performs well collectively. As such, program graphs incrementally grow and break apart at their root node, i.e. from the top up/down. While the team and program population sizes vary throughout evolution, the number of root teams to maintain in the population is a function of the number of tasks and the $nElite$ parameter (See step \ref{ga-sel}). Whenever a root team is subsumed within a program graph, it will first be cloned and the \textit{clone} becomes the internal node (See Figure \ref{fig:tpg}). Thus, as hierarchies grow, they must directly compete with their (simpler) subgraphs (i.e. prior to the addition of a new root node). This clone-when-subsumed constraint ensures that root teams with strong performance are not subsumed within a weaker-performing program graph. Without cloning, the subsumed root behaviour would no longer be part of the pool of independent agents, and its (high-fitness) stand-alone behaviour would be lost until the hierarchy breaks down. 


In summary, the hierarchical complexity and interdependency between teams in program graphs emerges entirely through interaction with the task environment.  As a program graph operates, the subset of teams/programs that require execution is dynamically selected at run-time based on the current input sample and the content of stateful memory. This has two important implications: 1) Teams are free to specialize on particular aspects of the problem and may be switched in and out of the model as needed; and 2) Program graphs can dynamically select inputs and stateful memory registers that are relevant to the current state observation (i.e. inputs and memory registers indexed by programs along the active path) while ignoring inputs/memories that are not important at the current point in time. This is conceptually similar to the modular structures and \textit{attention} mechanisms explored by Goyal at. al. \cite{goyal2019recurrent}, in which these properties were shown to improve generalization and in dynamic memory problems. However, in that case the total number of "modules" per solution required prior specification, as did the number of "active" modules at any point in time. In this work we are specifically interested in how these model characteristics emerge through compositional evolution. Section \ref{sec:eval} will demonstrate how these properties support hierarchical task decomposition in multi-task reinforcement learning.

\section{Training and Test Performance}\label{sec:eval}

\begin{figure}[!htb]
	\centering
	\subfigure[CartPole]{\includegraphics[height=3.7cm]{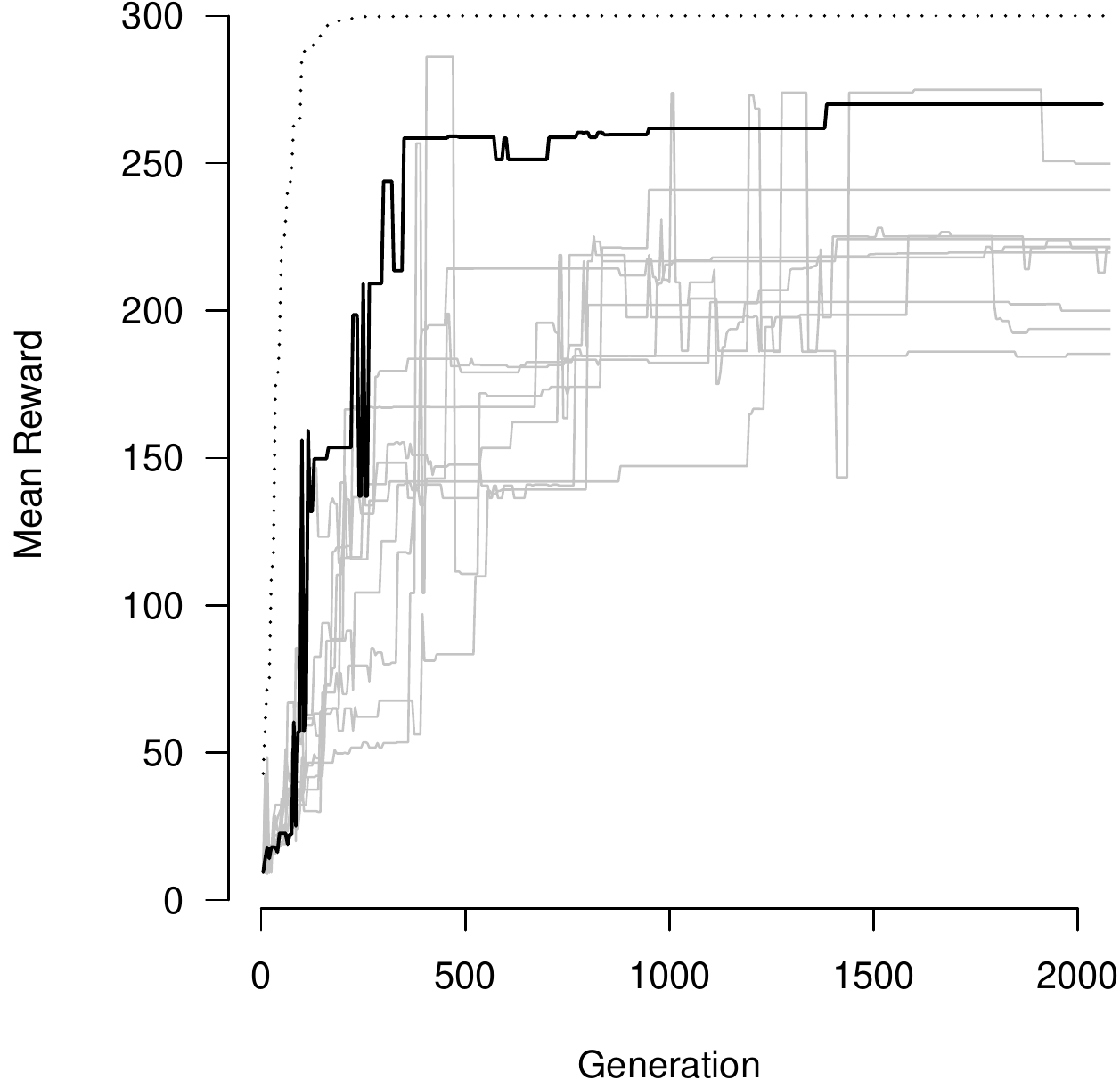}\label{fig:train-test-tsk-1}}
	\hspace{0.1cm}
	\subfigure[Acrobot]{\includegraphics[height=3.7cm]{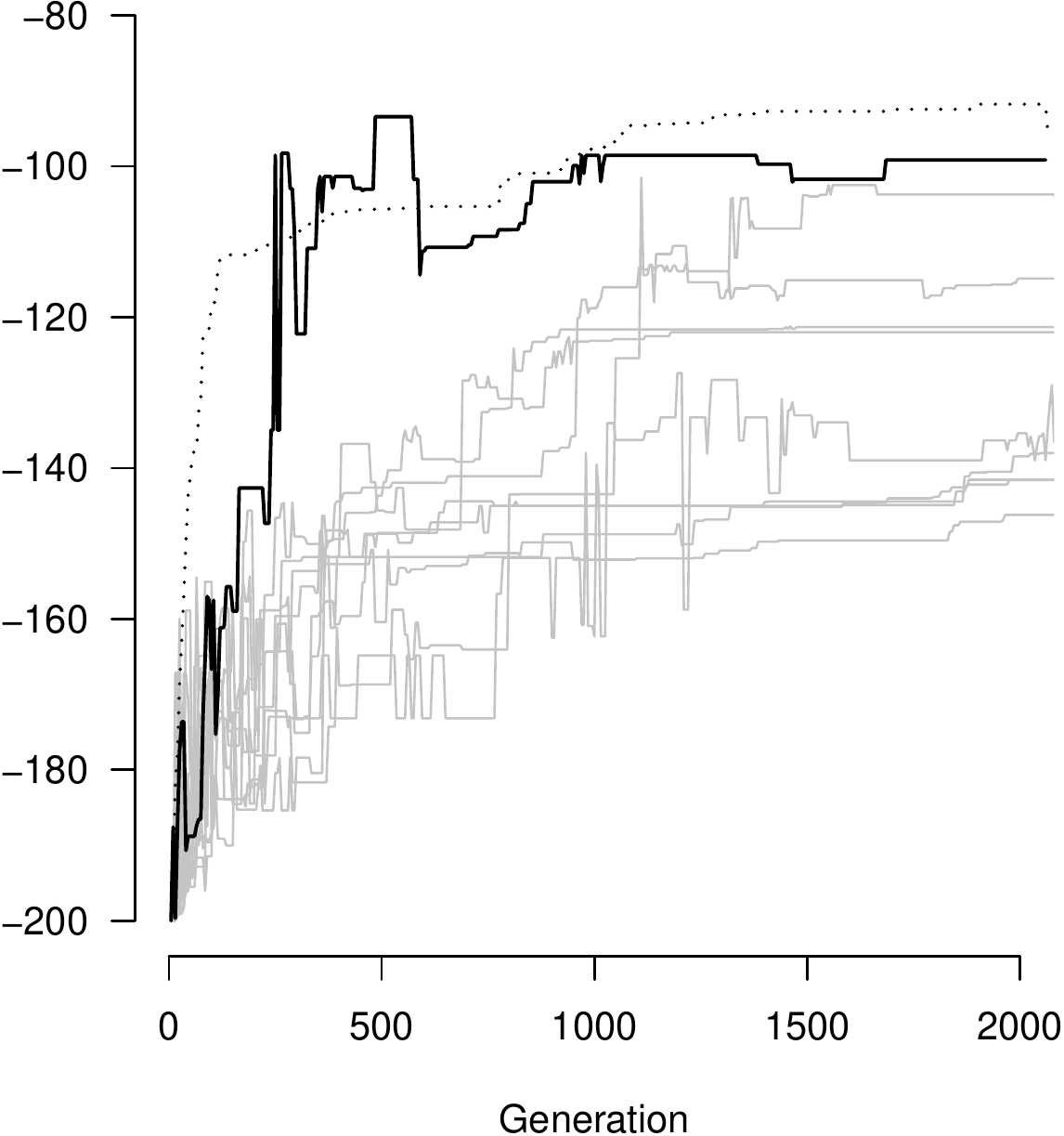}\label{fig:train-test-tsk-2}}
	\hspace{0.1cm}
	\subfigure[CartCentering]{\includegraphics[height=3.7cm]{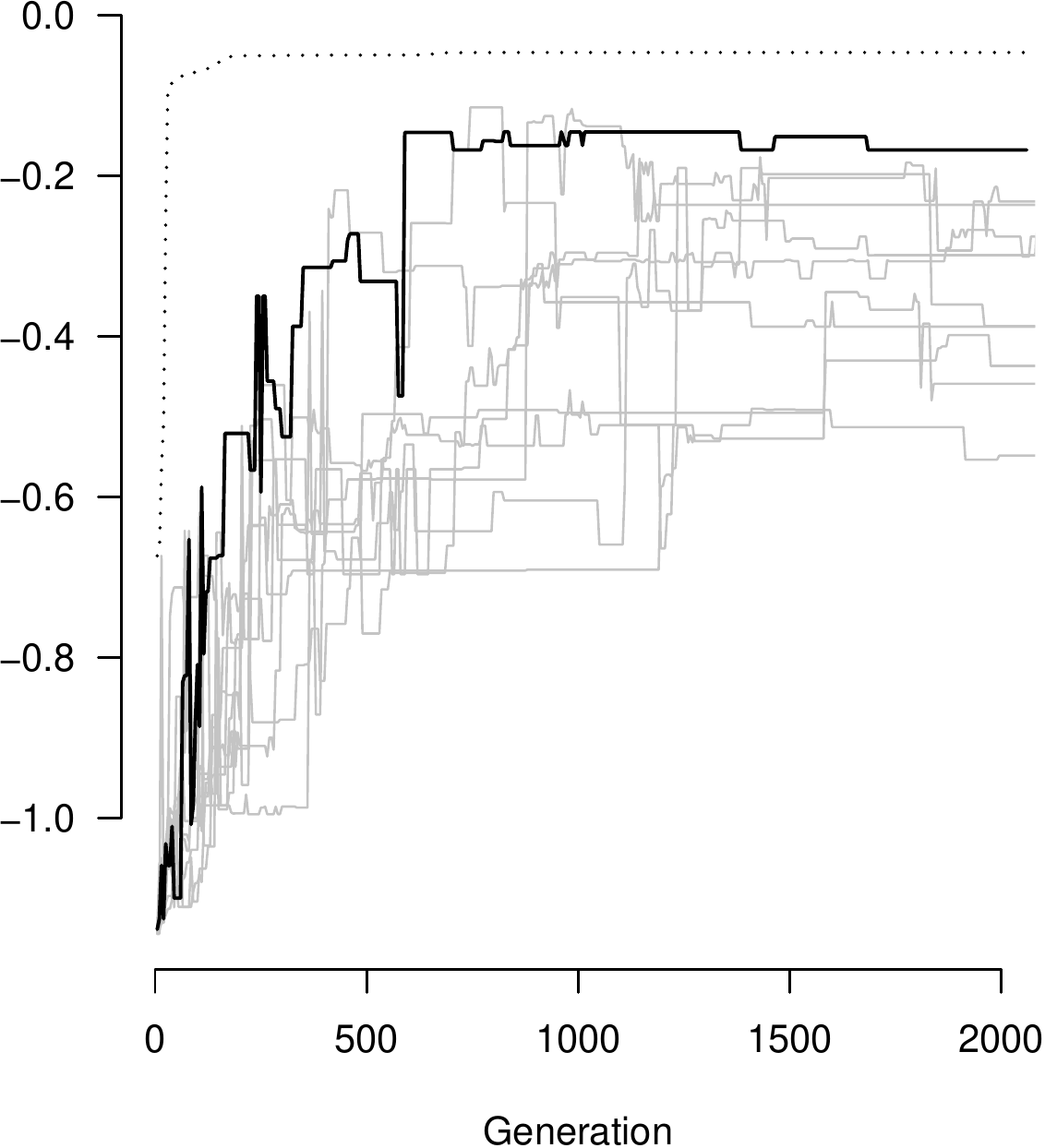}\label{fig:train-test-tsk-3}}
	\subfigure[Pendulum]{\includegraphics[height=3.7cm]{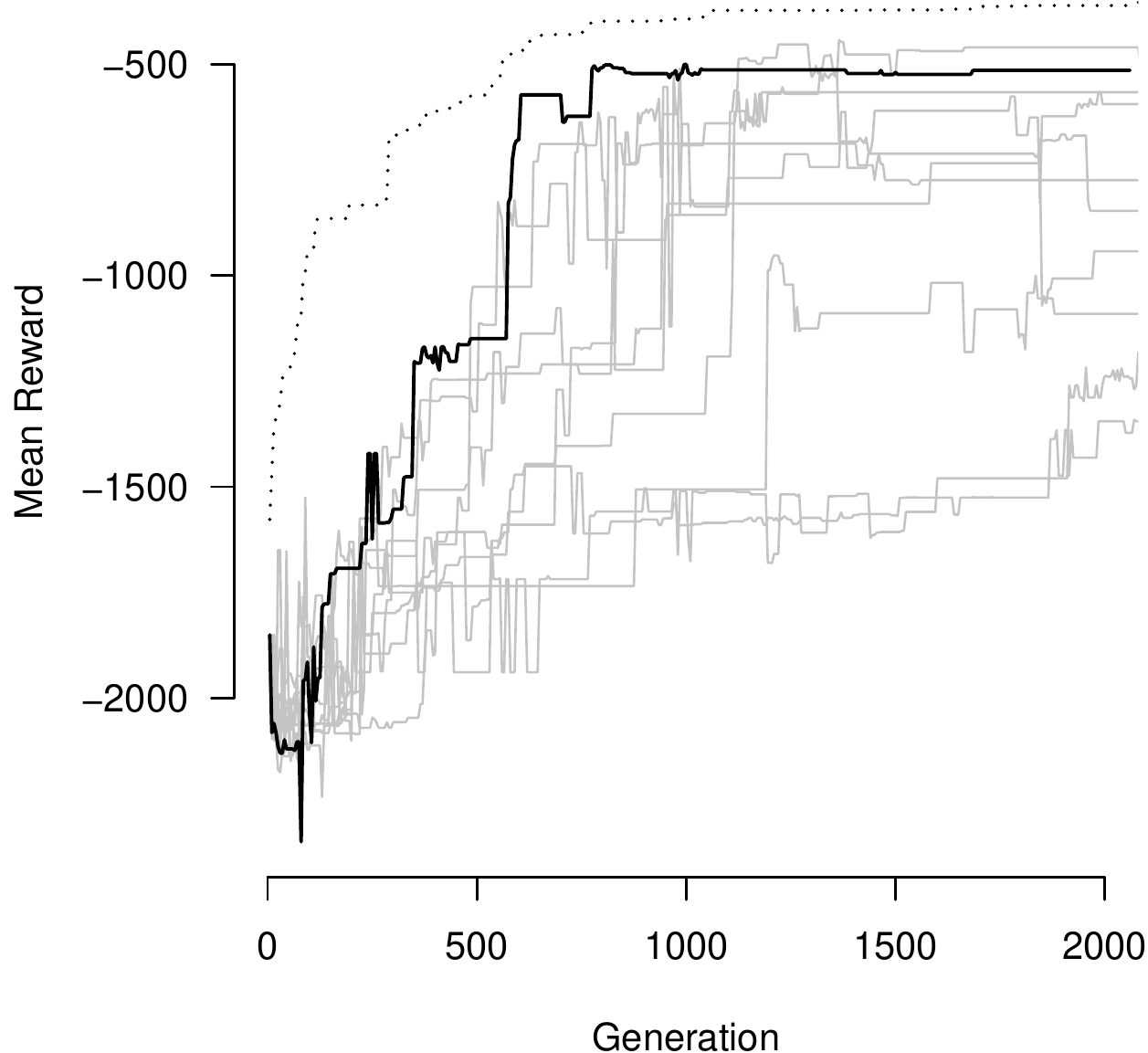}\label{fig:train-test-tsk-4}}
	\hspace{0.1cm}
	\subfigure[MountainCar]{\includegraphics[height=3.7cm]{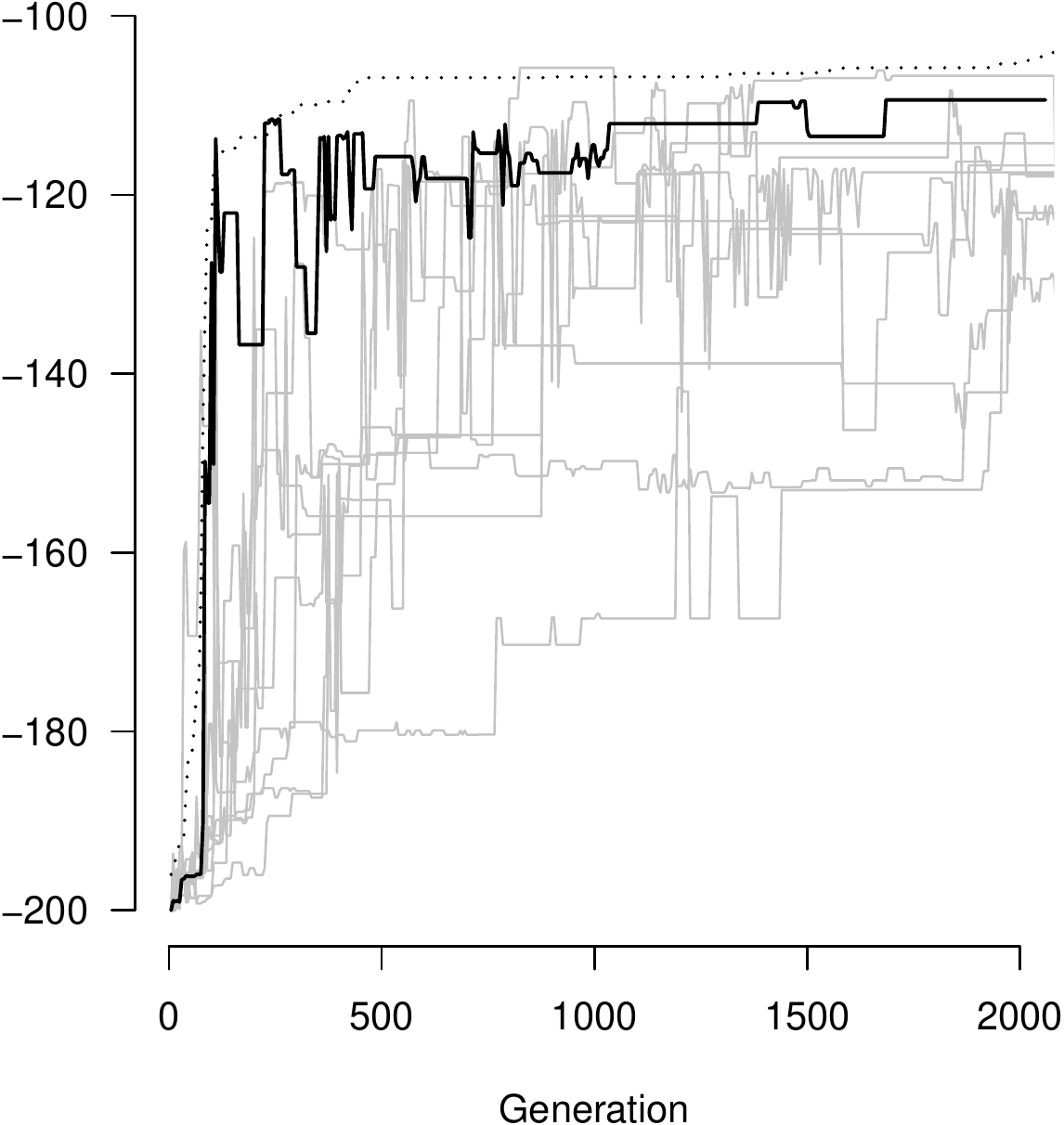}\label{fig:train-test-tsk-5}}
	\hspace{0.1cm}
	\subfigure[MountainCarC.]{\includegraphics[height=3.7cm]{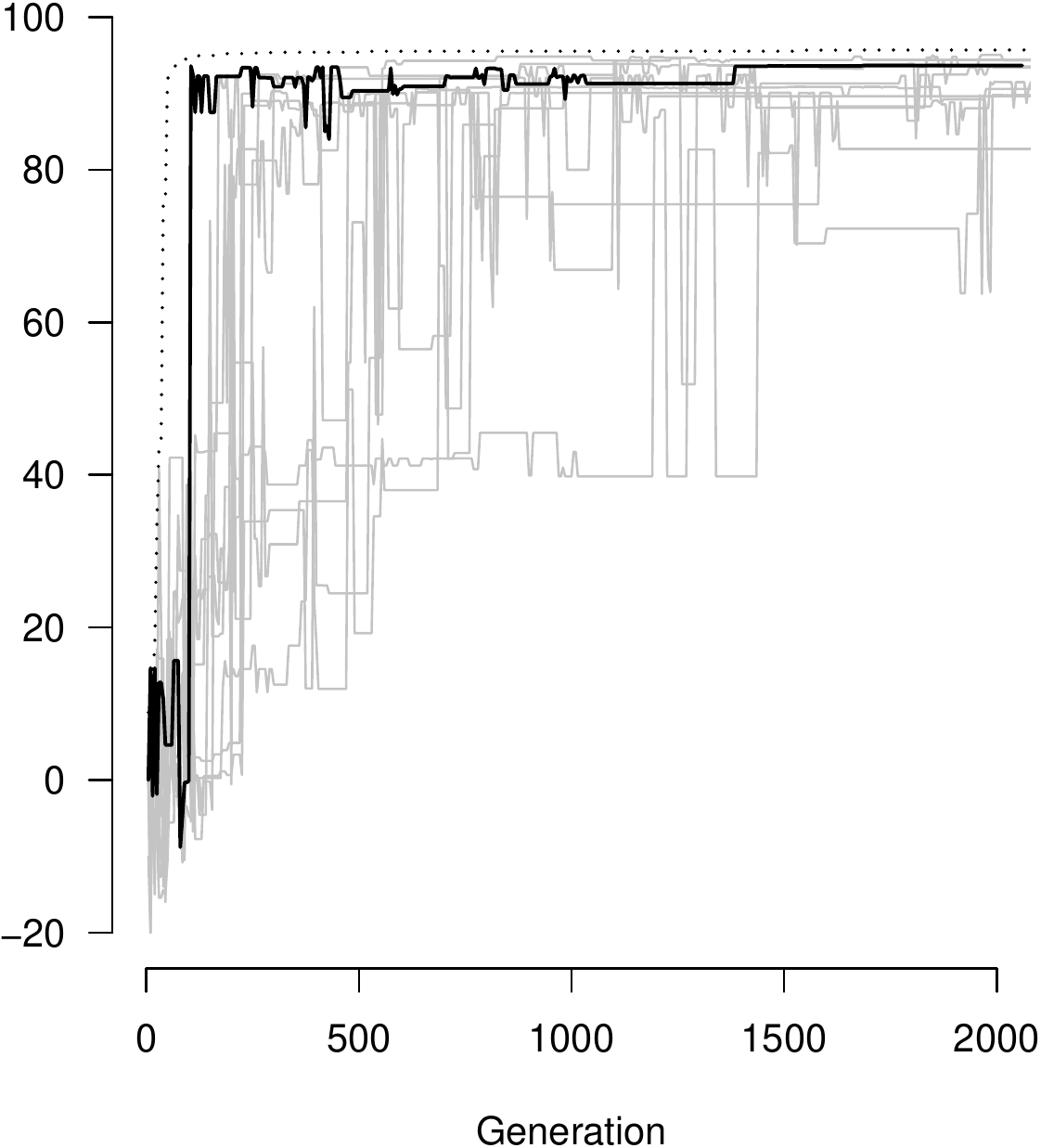}\label{fig:train-test-tsk-6}}
	\caption{Summarized TPG learning curves over 10 independent \ac{MTRL} runs. Rewards are averaged over 100 episodes with random initial conditions. Dotted line represents median test reward of best single-task program graphs (i.e. each plot reports median (over 10 runs) reward for the unique single-task champion identified for each task). Solid lines represent fitness of best multi-task program graph for each run, with the best over all runs in black (i.e. black line in each plot represents performance of the \textit{same} multi-task graph).}
	\label{fig:train-test}
\end{figure}

Figure \ref{fig:train-test} provides a summary of multi-task TPG learning curves over 10 independent runs. At intervals of 5 generations, the program graph with the highest training reward is identified for each task set (i.e. 63 unique sets, see Section \ref{ga-sel}), and this agent is evaluated in 100 test episodes for each task. Each test episode begins with random initial conditions not seen during training. Figures \ref{fig:train-test-tsk-1} to \ref{fig:train-test-tsk-6} report the average test reward at every 5-generation test interval. A dotted line represents the median of champion single-task program graphs (i.e. each plot reports median mean reward for the unique single-task champion identified for each task, at each test interval).  Single-task scores provide a benchmark for task difficulty. Some, but not all, tasks have a score threshold indicating when the task is considered solved. For example, CartPole is considered solved if the agent can balance the pole for an average of 195 timesteps over 100 episodes, which corresponds to a reward of 195 in Figure \ref{fig:train-test}. Within $\approx 500$ generations, TPG single-task scores (dotted line) reach a quality of behaviour in which all tasks can reasonably be considered solved. Section \ref{sec:leaderboard} makes a direct comparison with state-of-the-art single-task behaviours. 

\begin{figure}[!htb]
	\centering
	\includegraphics[width=6cm]{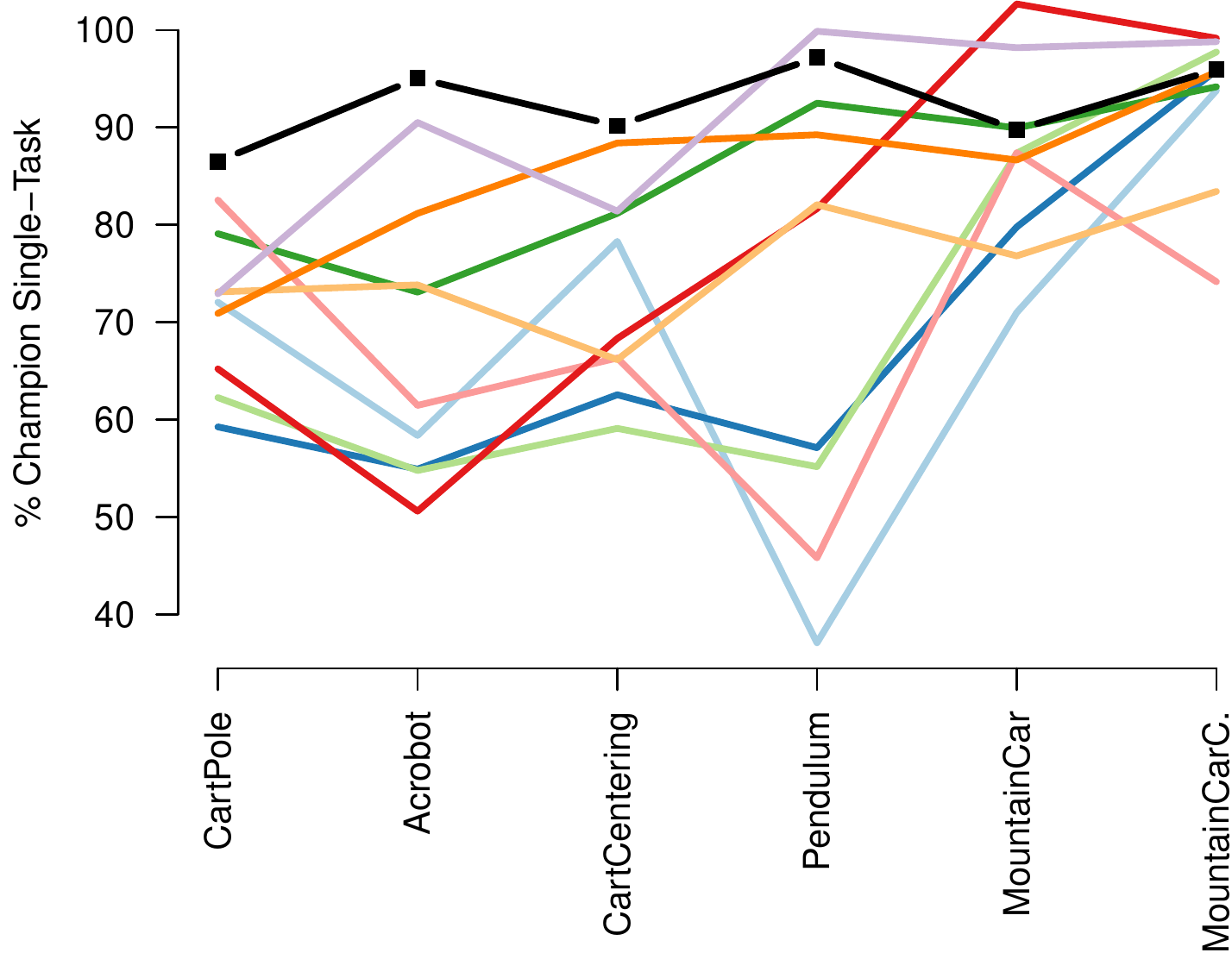}
	\caption{Comparison of multi-task agent test scores over 10 independent runs, normalized by the score of the best single-task agent in each task. Normalized score for multi-task agent $a_i$ in task $t_j$ is calculated as $(sc(a_i,t_j) - sc_{rand}(t_j))/(sc(st_{max}(t_j)) - sc_{rand}(t_j)))$ where $sc(a_i,t_j)$ is the mean score for agent $a_i$ in task $t_j$, $sc_{rand}(t_j)$ is the mean score for an agent that takes random actions in task $t_j$, and $sc(st_{max}(t_j))$ is the max single-task score in task $t_j$.}
	\label{fig:multi-run-test}
\end{figure}

Solid lines in Figure \ref{fig:train-test} represent average test reward of the best multi-task program graph for each run. At each test interval, the multi-task champions identified in each run may exhibit a unique performance trade-off over the 6 tasks. As such, it is not informative to report the average or median multi-task score over multiple runs. Thus, Figure \ref{fig:train-test} reports multi-task scores for each run individually (grey lines), with the best over all runs in black. That is, the black line in each task plot represents test reward of the \textit{same} multi-task graph at each 5-generation interval. By generation $\approx1000$, the single best multi-task program graph is competent in all 6 tasks. Scores for this champion are compared to single-task TPG scores in Figure \ref{fig:multi-run-test}, along with the champion multi-task scores from the 9 other TPG runs. It is apparent that the best run produced a multi-task agent that reaches roughly $90\%$ of the best single-task agent scores in all 6 tasks (black line), while 5/10 runs produced multi-task agents that reached at least $60\%$ of the single-task scores.

\subsection{Comparison with Fully-Observable Single-Task Leaderboard}\label{sec:leaderboard}

\begin{figure}[!htb]
	\centering
	\includegraphics[width=6cm]{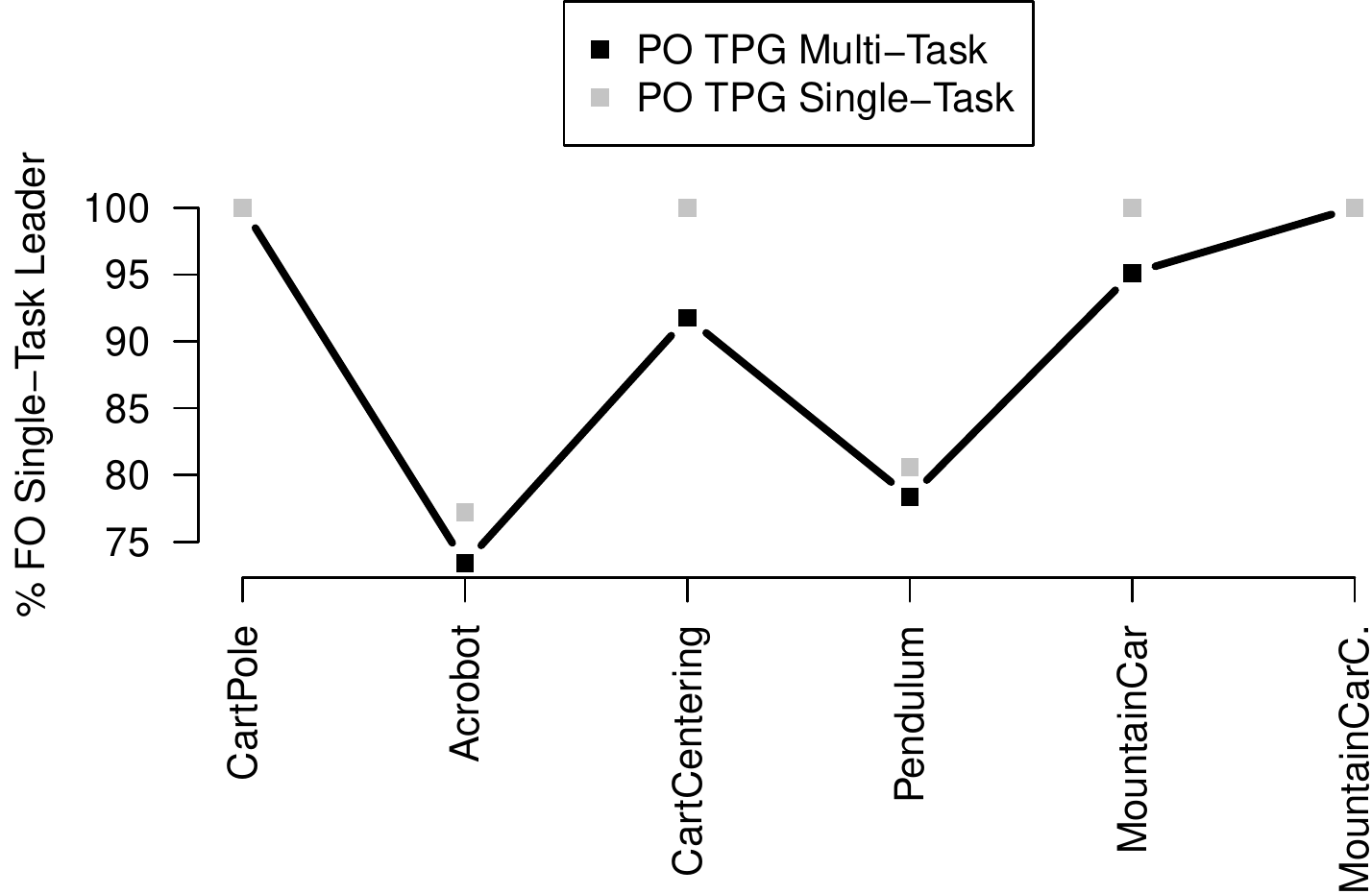}
	\caption{Comparison of multi-task and single-task TPG agent test scores, normalized by the score of the best agent from OpenAI's leaderboard at \href{https://github.com/openai/gym/wiki/Leaderboard}{https://github.com/openai/gym/wiki/Leaderboard}. Note that all leaderboard agents were trained independently for each task in fully-observable versions of the environment. Normalized score for multi-task agent $a_i$ in task $t_j$ is calculated as $(sc(a_i,t_j) - sc_{rand}(t_j))/(sc(st_{max}(t_j)) - sc_{rand}(t_j)))$ where $sc(a_i,t_j)$ is the mean score for TPG agent $a_i$ in task $t_j$, $sc_{rand}(t_j)$ is the mean score for an agent that takes random actions in task $t_j$, and $sc(st_{max}(t_j))$ is the best score on OpenAI's leaderboard \textit{with} an accompanying writeup at the time of this writing. In the case of tasks with a threshold over which they are considered solved (CartPole, both version of Mountain Car), this threshold is used as $sc(st_{max}(t_j))$. CartCentering is not yet part of OpenAI Gym but the time-optimal control program for fully observable state is known \cite{Koza1992}, thus this time-optimal controller is used as $sc(st_{max}(t_j))$. Sources for Acrobot and Pendulum leaders are from the Distributed Distributional Deep Deterministic Policy Gradient algorithm, D4PG \cite{barthmaron2018distributed}.}
	\label{fig:leaderboard}
\end{figure}

OpenAi Gym's leaderboard provides a repository to track and compare \ac{RL} algorithms. Figure \ref{fig:leaderboard} compares the performance of multi-task and single-task TPG in partially-observable classic control environments with the best scores in the leaderboard. Note that all leaderboard agents were trained and tested independently for each task with \ac{FO} versions of the environments. Multi-Task learning in \ac{PO} environments is a significantly more challenging problem. The champion Multi-task TPG agent, trained and tested in \ac{PO} environments, reaches at least $90\%$ of the best leaderboard score in 4/6 tasks, and $\approx 80\%$ and $\approx 75\%$ in the remaining two.   While the Multi-Task TPG agent does not quite match the leaderboard scores, it reaches a general quality of behaviour in which all tasks can be considered solved. Section \ref{sec:replay} provides a detailed analysis of the structure and behaviour of the champion \ac{TPG} \ac{MTRL} agent.

\subsection{Ablation Study}\label{sec:ablation}

\begin{figure}[!htb]
	\centering
	\includegraphics[width=6cm]{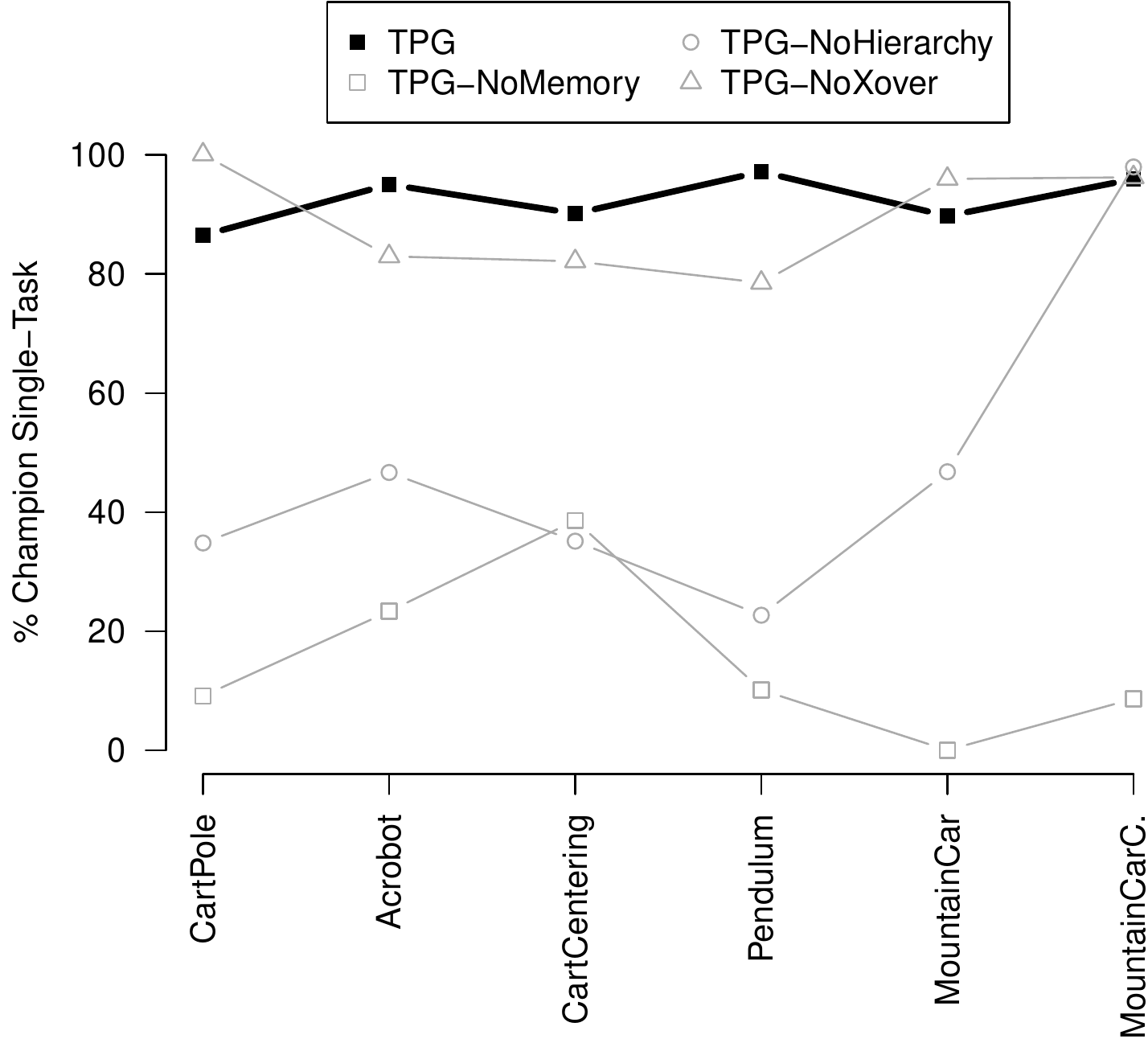}
	\caption{Multi-Task Ablation. Plot provides normalized test scores for the single best multi-task program graph discovered when critical components of the TPG algorithm are removed.  Normalized score for multi-task agent $a_i$ in task $t_j$ is calculated as $(sc(a_i,t_j) - sc_{rand}(t_j))/(sc(st_{max}(t_j)) - sc_{rand}(t_j)))$ where $sc(a_i,t_j)$ is the mean score for agent $a_i$ in task $t_j$, $sc_{rand}(t_j)$ is the mean score for an agent that takes random actions in task $t_j$, and $sc(st_{max}(t_j))$ is the max single-task score in task $t_j$.}
	\label{fig:ablation}
\end{figure}

\begin{figure}[!htb]
	\centering
	\includegraphics[width=1\textwidth]{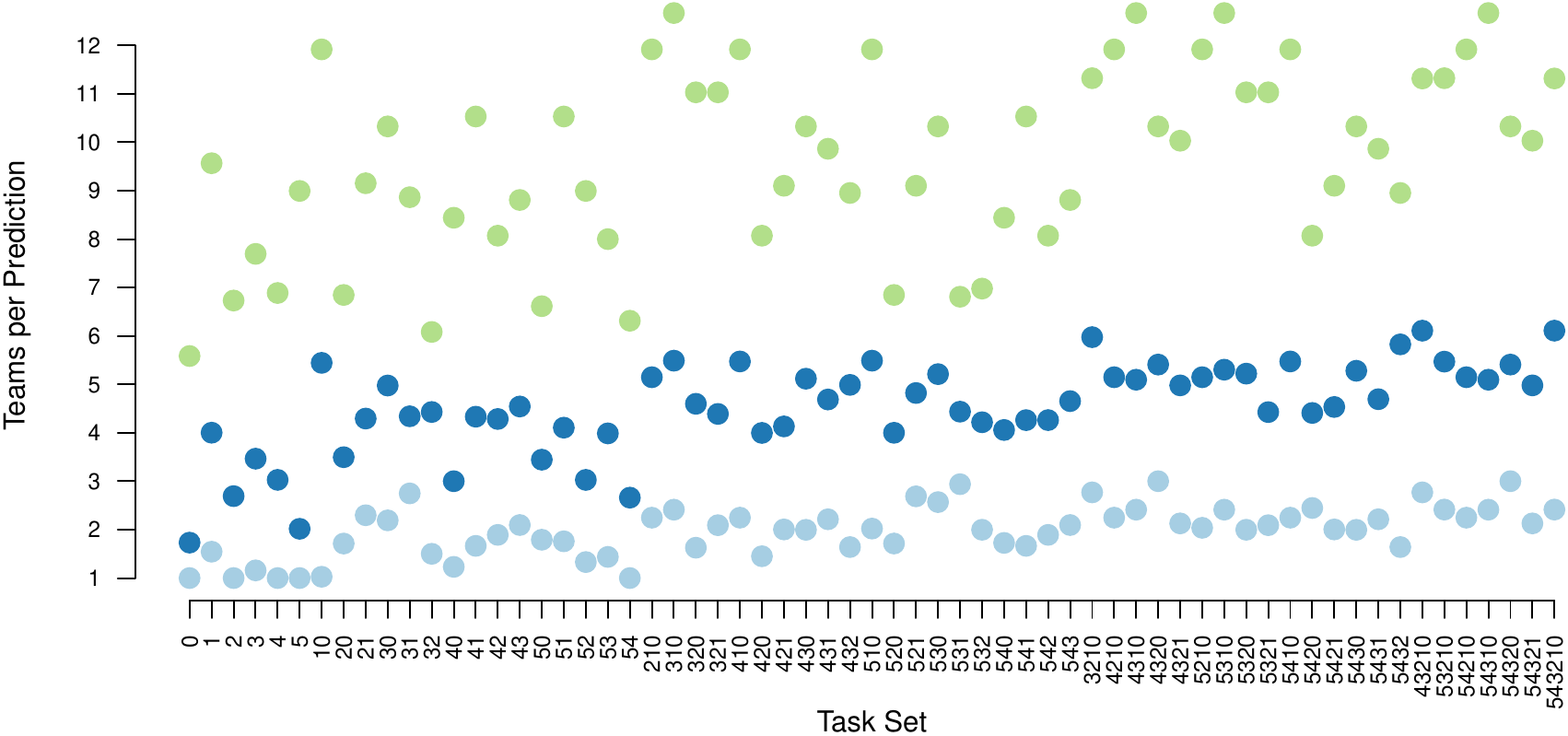}
	\caption{Hierarchical complexity of the best program graph discovered for each set in the task power set, as measured by the average number of teams visited per timestep (prediction) over 100 test episodes. Points indicate max, median, and min over 10 independent runs. Tasks are numbered in the following order: 0-CartPole, 1-Acrobot, 2-CartCentering, 3-Pendulum, 4-MountainCar, 5-MountainContinuous}
	\label{fig:taskSet-v-hierarchy}
\end{figure}

In order to confirm the significance of critical components of the TPG algorithm (Section \ref{sec:tpg}), an ablation study is performed with 3 additional experiments, each with one component removed. Figure \ref{fig:ablation} summarizes the ablation results. For clarity, we limit the ablation analysis to a comparison of the single best multi-task agent produced from each experiment, as identified by the multi-task selection procedure described in Section \ref{ga-sel}. Without crossover (TPG-NoXover), the best multi-task agent still achieves at least $80\%$ of single-task performance in all tasks. Compared to full TPG, TPG-NoXover is equal in one task (MountainCarContinuous), better in 2 tasks, and worse in 3 tasks. Also, its single worst normalized score (in Pendulum) is less that any score from TPG. As such, it would be ranked behind TPG by the multi-task ranking procedure outlined in Section \ref{ga-sel}. TPG-NoMemory refers to the scenario in which all registers (internal and shared) are \textit{stateless}. That is, registers are reset to zero prior to each program execution. In this case, agents have no means of building an internal model of the environment and integrating state information across timesteps during an episode, something that is required in partially-observable environments. As a result, the best TPG-NoMemory agent is weak, achieving well below $50\%$ of single-task agent scores in all tasks. Finally, TPG-NoHierarchy refers to the experiment in which TPG is parameterized with $p_{atomic}=1.0$. In this case, TPG's ability to construct program graphs is disabled, and all evolved agents will take the form of a single team of programs. As described in Section \ref{sec:tpg}, TPG supports multi-task operation by automatically decomposing the overall problem within the program graph hierarchy. In short, each team in the hierarchy is free to specialize on particular aspects of the overall multi-task problem, and the agent (program graph) is able to generalize by recombining various specialized team behaviours as it encounters different environmental scenarios over time. As seen in Figure \ref{fig:ablation}, when hierarchical development is disabled, the best multi-task agents can still specialize well in one environment (MountainCarContinuous, in this case), but are unable to generalize to other tasks. The importance of hierarchical task decomposition in multi-task learning is further evident in Figure \ref{fig:taskSet-v-hierarchy}, which reports the hierarchical complexity of decision-making (i.e. average number of teams visited per graph traversal during test) for the best agent in each combination of tasks in the task power set (See Section \ref{ga-sel}). While there is significant variation in hierarchical complexity, the larger task sets typically require agents which have subsumed more independent teams within their structure, and are thus able to generalize across a wider range of environments. The next section will examine structural and behavioural properties of the best 6-task program graph.   

\section{Structure and Behaviour of Best Program Graph}\label{sec:replay}

\begin{figure}[!htb]
	\centering
	\subfigure{\includegraphics[width=.7\textwidth]{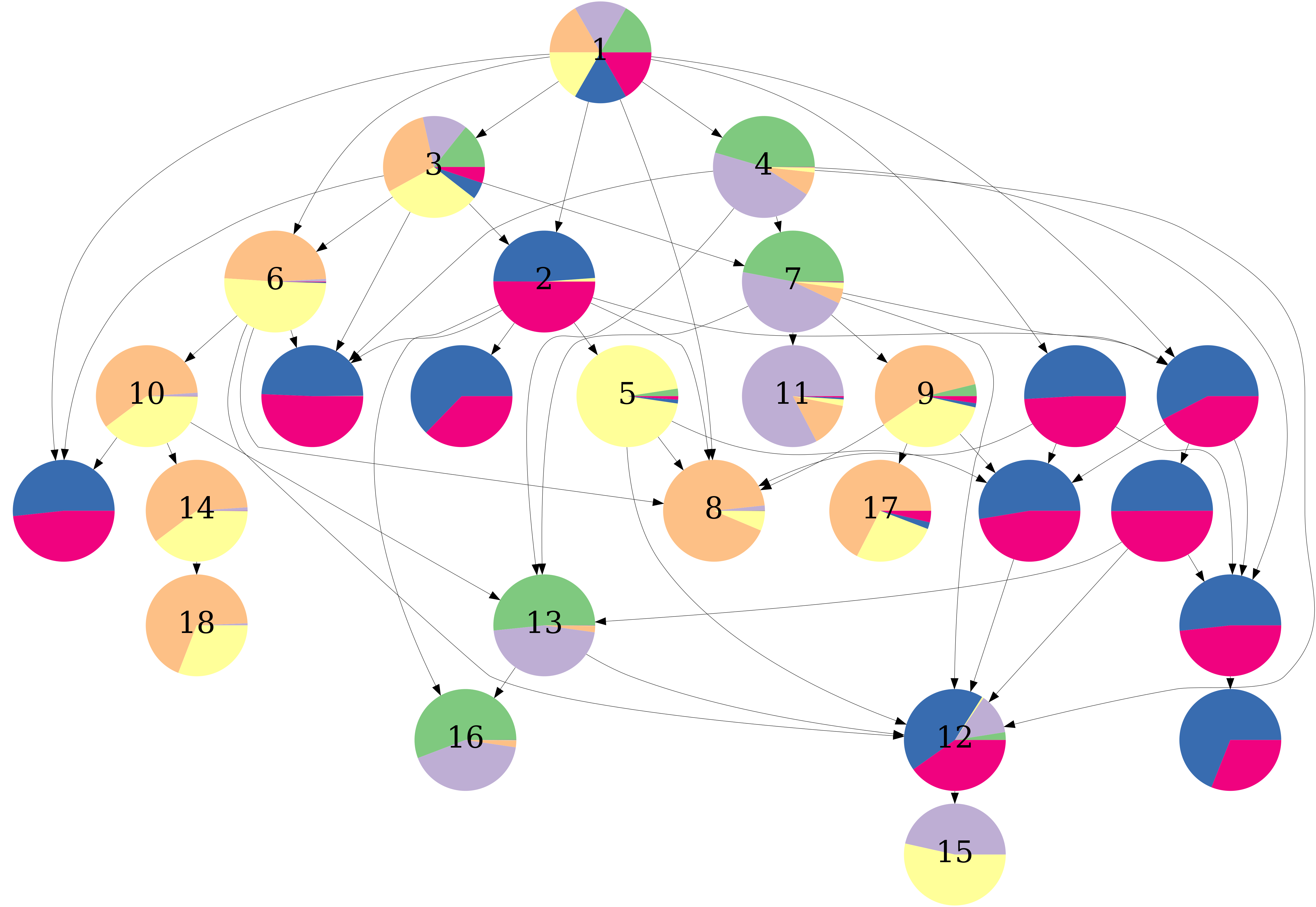}\label{fig:mtrl-graph}}
	\subfigure{\includegraphics[width=1\textwidth]{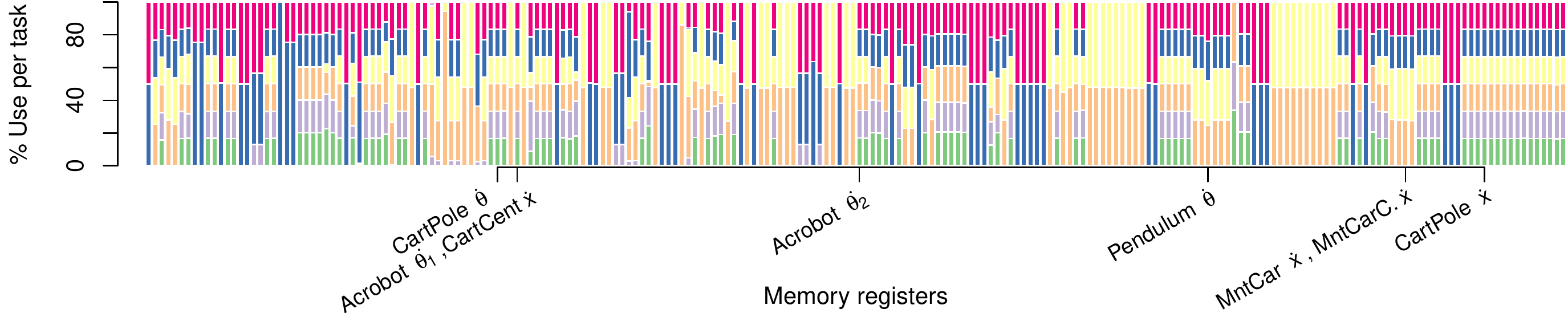}\label{fig:mem-reg}}
	\caption{Champion multi-task program graph. Each node represents one team of programs. Node charts illustrate proportion of timesteps in which each team was visited over 100 test episodes in each task. For example, the root node is visited in every timestep, thus proportions are equal for \colorbox{cartPoleCol}{CartPole}, \colorbox{acrobotCol}{Acrobot}, \colorbox{cartCenteringCol}{CartCentering}, \colorbox{pendulumCol}{Pendulum}, \colorbox{mountainCarCol}{MountainCar}, and \colorbox{mountainCarContinuousCol}{MountainCarContinuous}. Barplot shows proportion of per-task access (read or write) for all shared memory registers used by this program graph. Registers are distributed throughout graph but can be loosely tied to specific nodes by task decomposition. For example, registers with even proportions (Right-Hand Side of barplot) must be in root node. Node numbering and register x-axis labels are referenced in Section \ref{sec:replay}, \ref{sec:rtc}, and \ref{sec:vel-prediction} text.}
\end{figure}

Figure \ref{fig:mtrl-graph} illustrates the champion multi-task program identified from the TPG experiment (black lines in Figures \ref{fig:train-test}, \ref{fig:multi-run-test}, \ref{fig:leaderboard}, \ref{fig:ablation}). For clarity only the team hierarchy is shown, individual programs are omitted. Recall that in each timestep, graph traversal begins at the root node and follows one path through the graph until an leaf program is found. Since every team has at least one leaf program , graph traversal can terminate at any team. Each team is depicted by a pie chart indicating the proportion of timesteps in which it was visited over 100 test episodes in each task. Animations of this program graph interacting with all tasks are available here (Note hyperlinks, full addresses in \cite{cartPoleA,acrobotA,cartCenteringA,pendulumA,mountainCarCA}): \href{https://vimeo.com/547319808}{CartPole}, \href{https://vimeo.com/547319719}{Acrobot}, \href{https://vimeo.com/547319756}{CartCentering}, \href{https://vimeo.com/547319912}{Pendulum}, \href{https://vimeo.com/547319863}{MountainCarContinuous}. Animations depict the team hierarchy as well as individual programs. The active components in the graph are emphasized at each timestep, with the decision path (highest weight edges) highlighted in green. 



\begin{figure}[!htb]
	\centering
	\subfigure{\includegraphics[width=3.5cm]{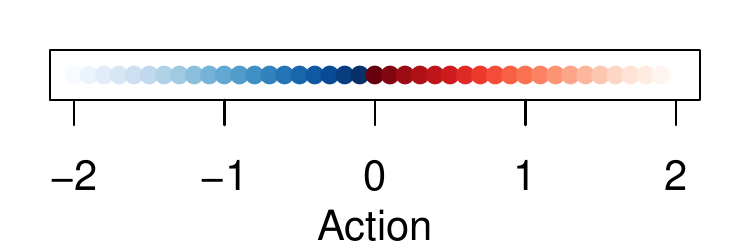}\label{fig:pendulum-action-legend}}\\
	\subfigure[Pendulum Task Decomposition]{\includegraphics[width=\textwidth]{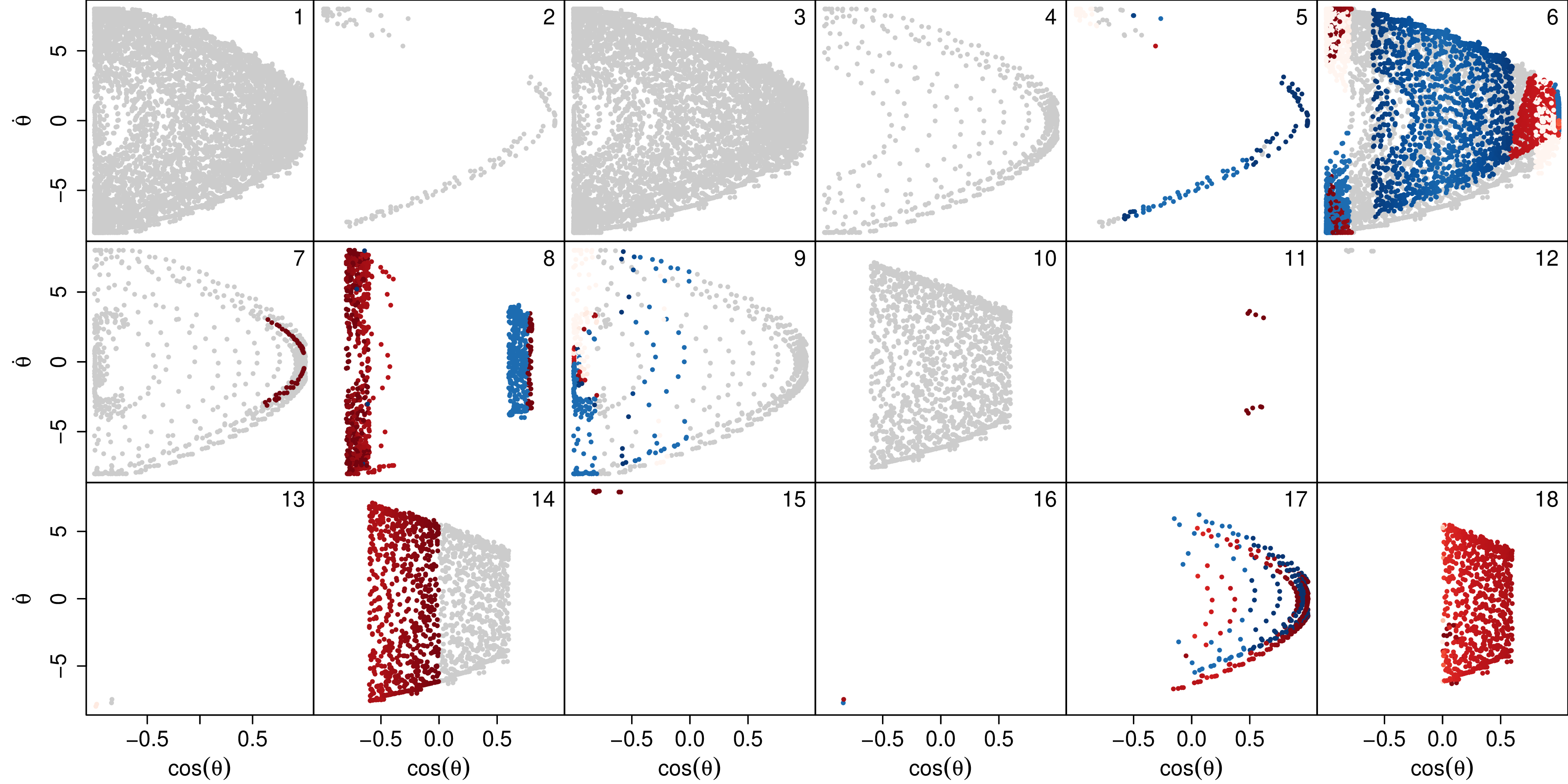}\label{fig:pend-decomp}}
	\subfigure{\includegraphics[width=1.5cm]{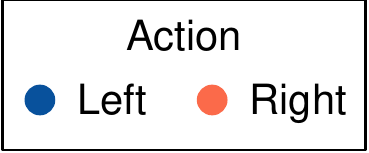}\label{fig:acrtCentering-action-legend}}\\
	\subfigure[CartCentering Task Decomposition]{\includegraphics[width=\textwidth]{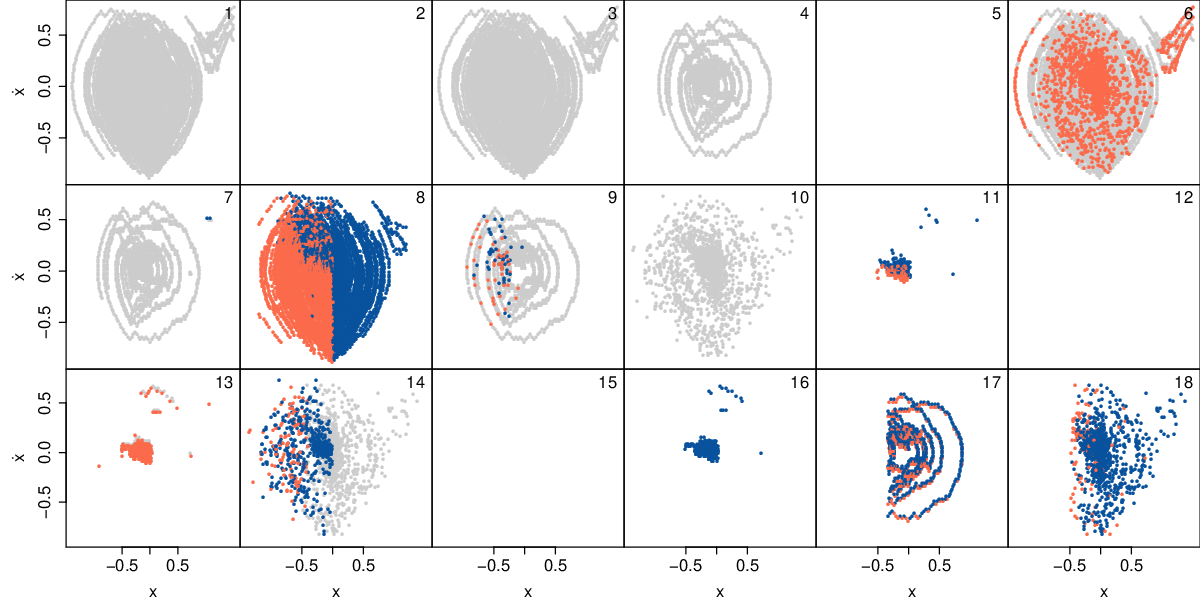}\label{fig:cartCentering-decomp}}
	\caption{Example task decompositions over 100 test episodes for the champion program graph depicted in Figure \ref{fig:mtrl-graph}. Each cell displays the points (in 2 dimensions of the problem state space) when each team was visited during graph execution. Colored dots indicate the team was the terminal stop and produced an atomic action, grey dots indicate the team was visited but ultimately passed execution to a lower-level team. Note that the vertical axis variable describes velocity of the system and is unobservable. Pendulum is a continous-action problem, while CartCentering is discrete-action. Color legends indicate color-coding of points with respect to actions.}
	\label{fig:tm-decomp}
\end{figure}

Naturally, MountainCar and MountainCarContinuous are closely related problems, thus it is not surprising that individual teams often generalize over these tasks. Similarly, teams often generalize over CartCentering and Pendulum, but the relationship between these tasks is less obvious. We can gain insight into how the team hierarchy behaves in these tasks by examining where each team is active within the system state space. The state of Pendulum and CartCentering can be fully described by two variables, only one of which is observable to the agent (See Table \ref{tbl:interface}). Each cell in Figure \ref{fig:tm-decomp} represents one numbered team in Figure \ref{fig:mtrl-graph}, and displays the points (in 2 dimensions of the state space) when the team was visited during graph execution. Each dot represents one timestep over 100 test episodes. Grey points indicated the team was visited at that step but ultimately passed execution to a lower-level team. Colored points indicate the team was the terminal stop and produced an atomic action at that timestep. For example, the root team (1) is active at every timestep but is never the terminal node. A common path through the graph for both tasks is [1, 3, 6, 10, 14, 18]. Notice that the behaviour of the terminal teams gets more specialized as execution moves down the hierarchy (colored dots are increasingly fewer and more concentrated). The hierarchy decomposes both tasks in this manner using the same path, and there are similarities in the nature of this decomposition. For example, see team/cell 14, which makes a clear distinction at $\approx 0$ in the observable state variable ($cos(\theta)$ in Pendulum and $x$ in CartCentering) before passing execution to team/cell 18. In other cases, for example team/cell 8, the behaviour of the terminal team decomposes these tasks entirely differently in the space of the observable variable. This indicates that the agent must be encoding some representation for the (unobservable) system velocity in memory and using this \textit{prediction} of velocity to determine the action. Finally, note that Pendulum is a continuous-action problem while CartCentering is discrete-action. It is clear that some teams are capable of providing actions for both cases (e.g. teams 6, 8, 14) while others specialize on one type of action (e.g. team 5). 

\subsection{Run-Time Complexity}\label{sec:rtc}

\begin{figure}[!htb]
	\centering
	\subfigure[CartPole]{\includegraphics[width=.485\textwidth]{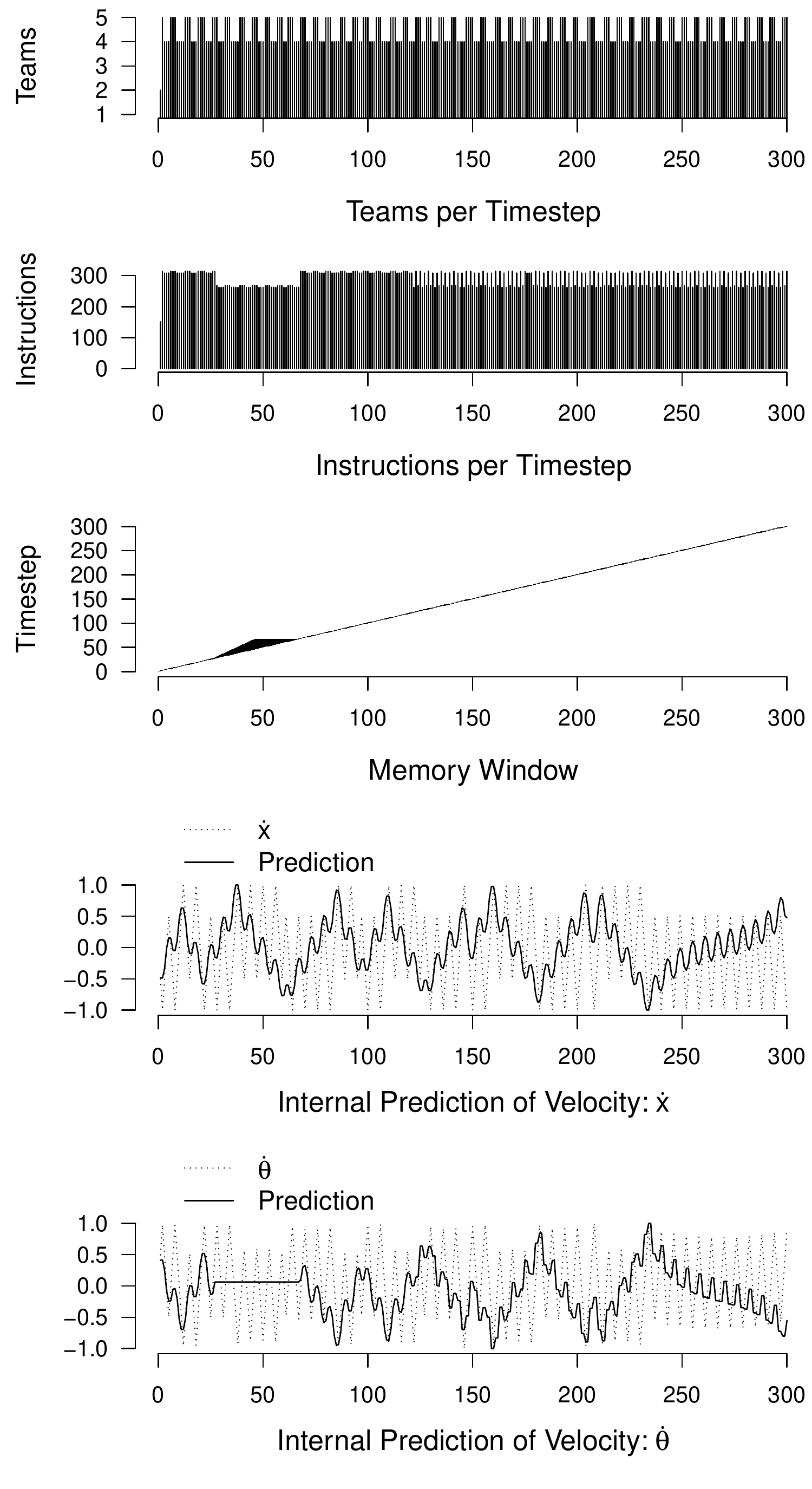}\label{fig:best_runTimeStats-tsk-1}}
	\hspace{0.05cm}
	\subfigure[Acrobot]{\includegraphics[width=.485\textwidth]{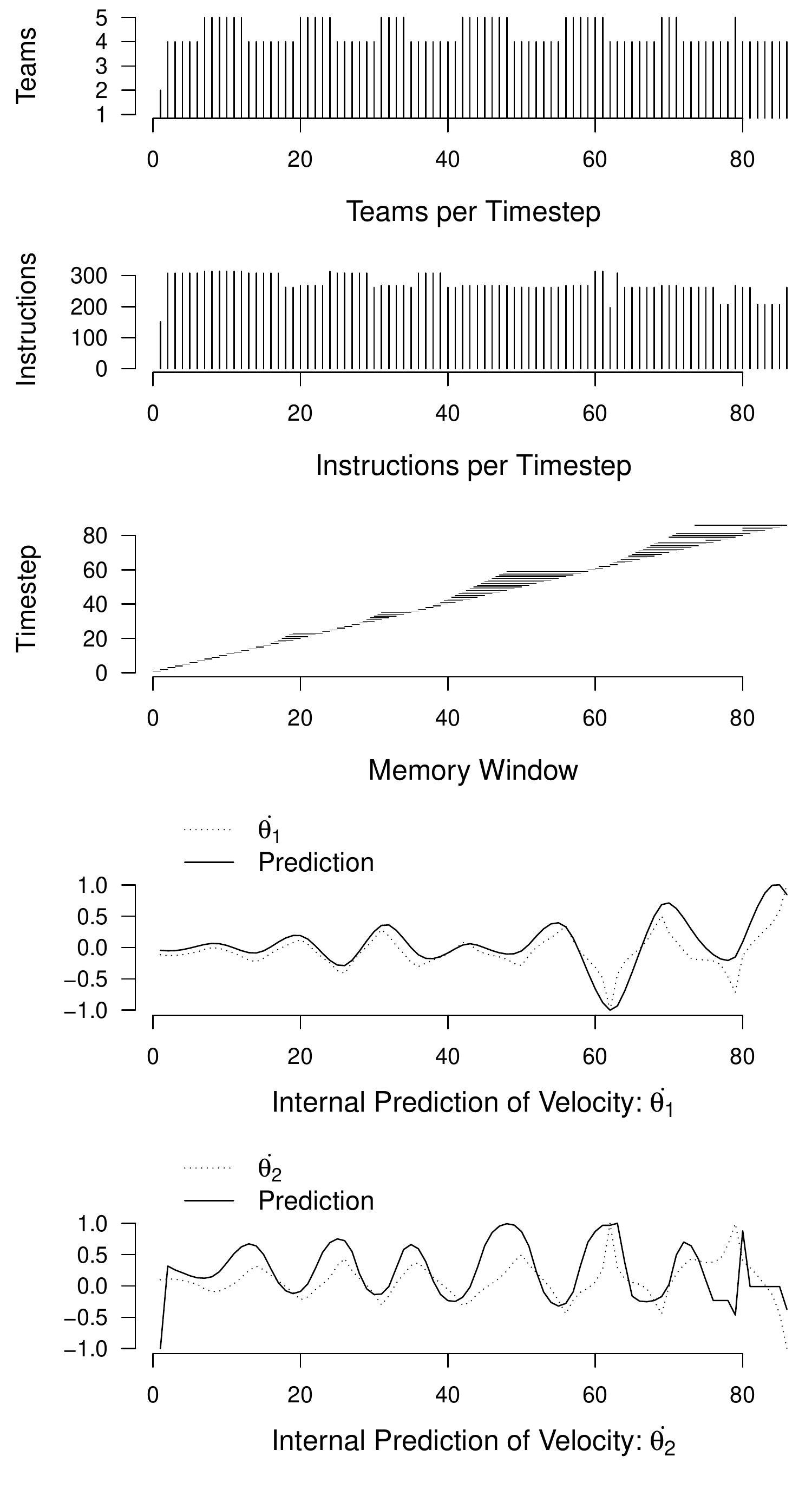}\label{fig:best_runTimeStats-tsk-2}}
	\caption{Time series data recorded during replay of best multi-task program graph (Figure \ref{fig:mtrl-graph}) under 1 episode in CartPole and Acrobot. x-axis is timesteps. See Sections \ref{sec:rtc}, \ref{sec:dynmem}, and \ref{sec:vel-prediction} for details. }
\label{fig:rtc-1}
\end{figure}

\begin{figure}[!htb]
	\centering
	\subfigure[CartCentering]{\includegraphics[width=.485\textwidth]{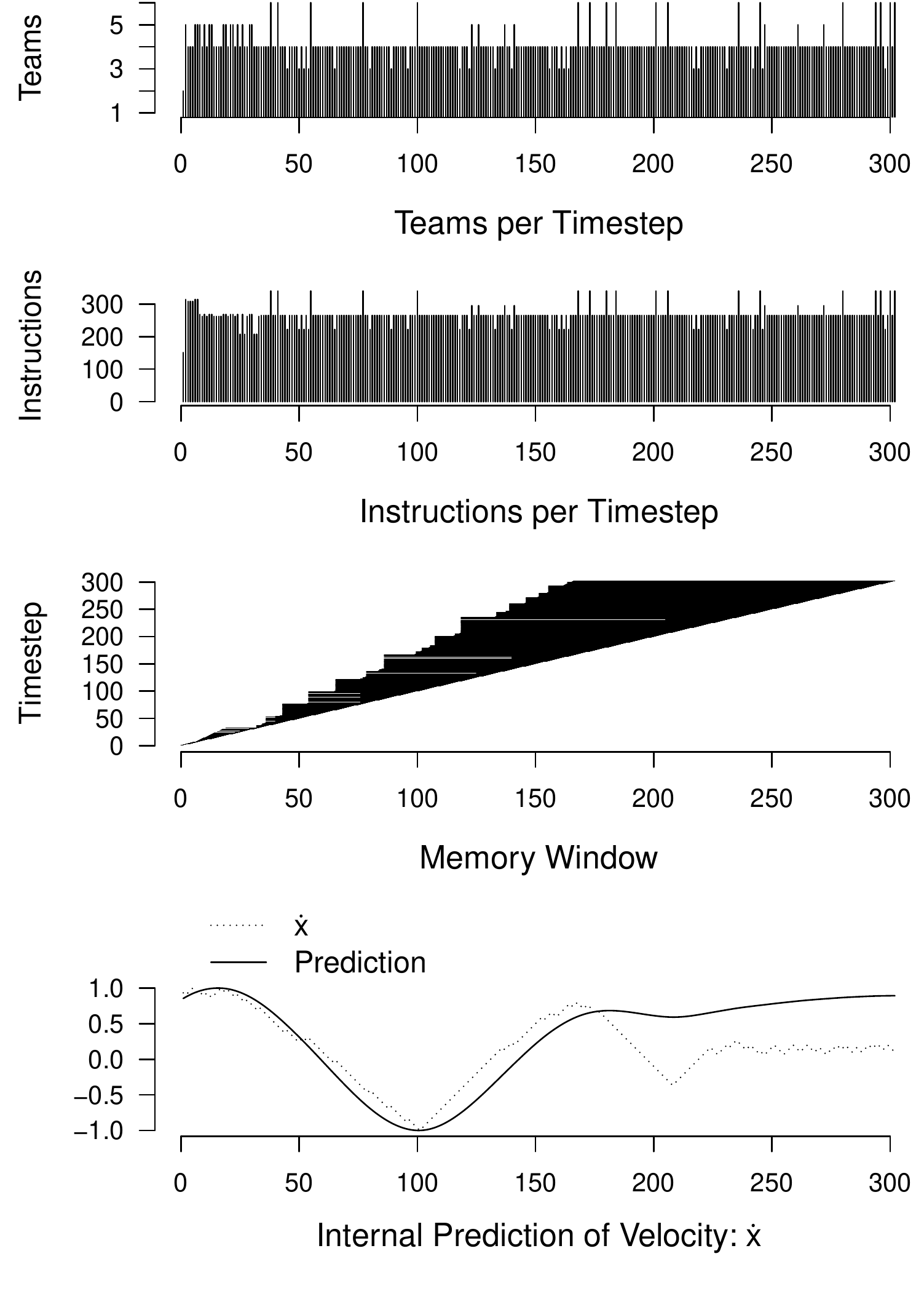}\label{fig:best_runTimeStats-tsk-3}}
	\hspace{0.05cm}
	\subfigure[Pendulum]{\includegraphics[width=.485\textwidth]{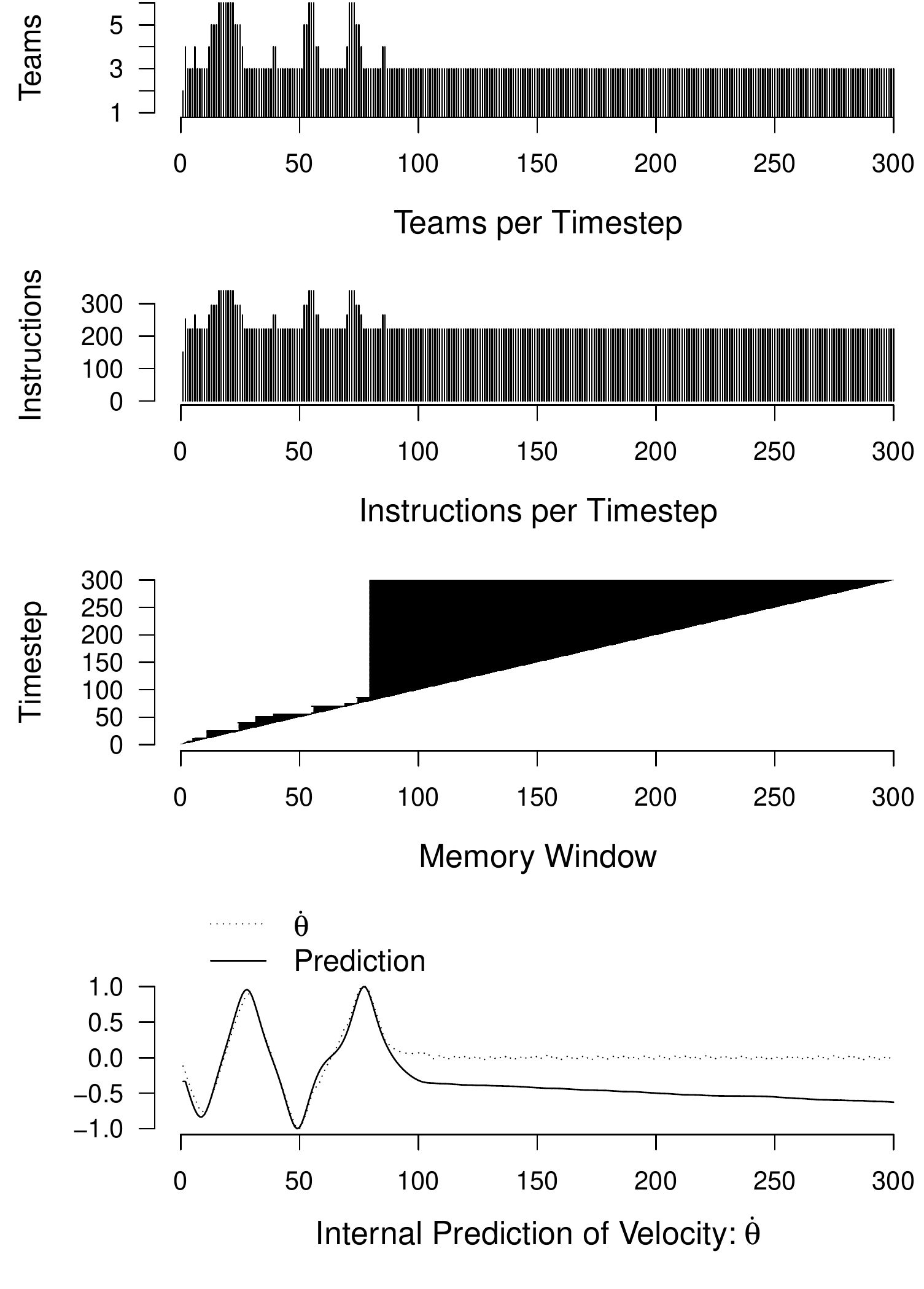}\label{fig:best_runTimeStats-tsk-4}}
	\subfigure[MountainCar]{\includegraphics[width=.485\textwidth]{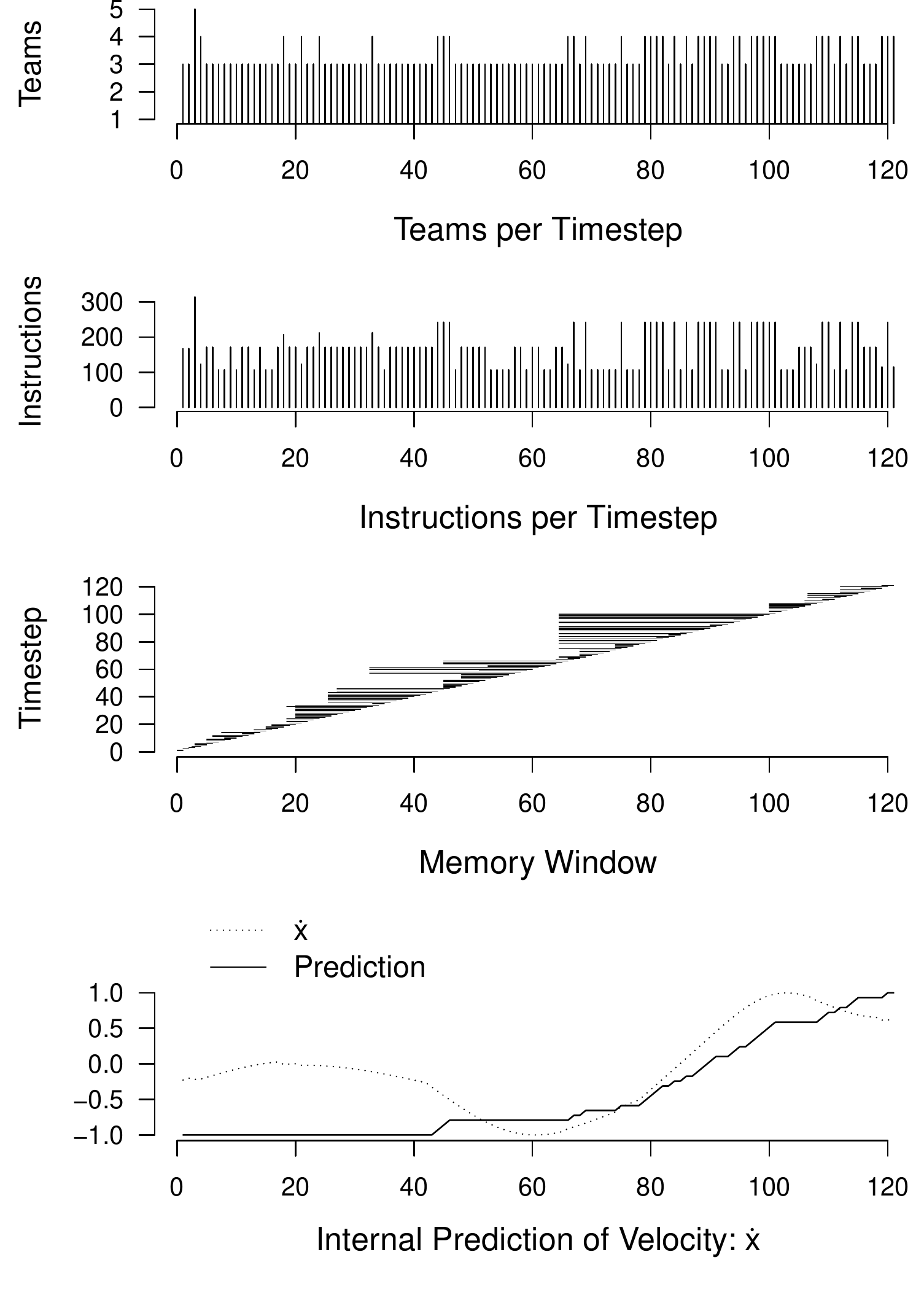}\label{fig:best_runTimeStats-tsk-5}}
	\hspace{0.05cm}
	\subfigure[MountainCarC.]{\includegraphics[width=.485\textwidth]{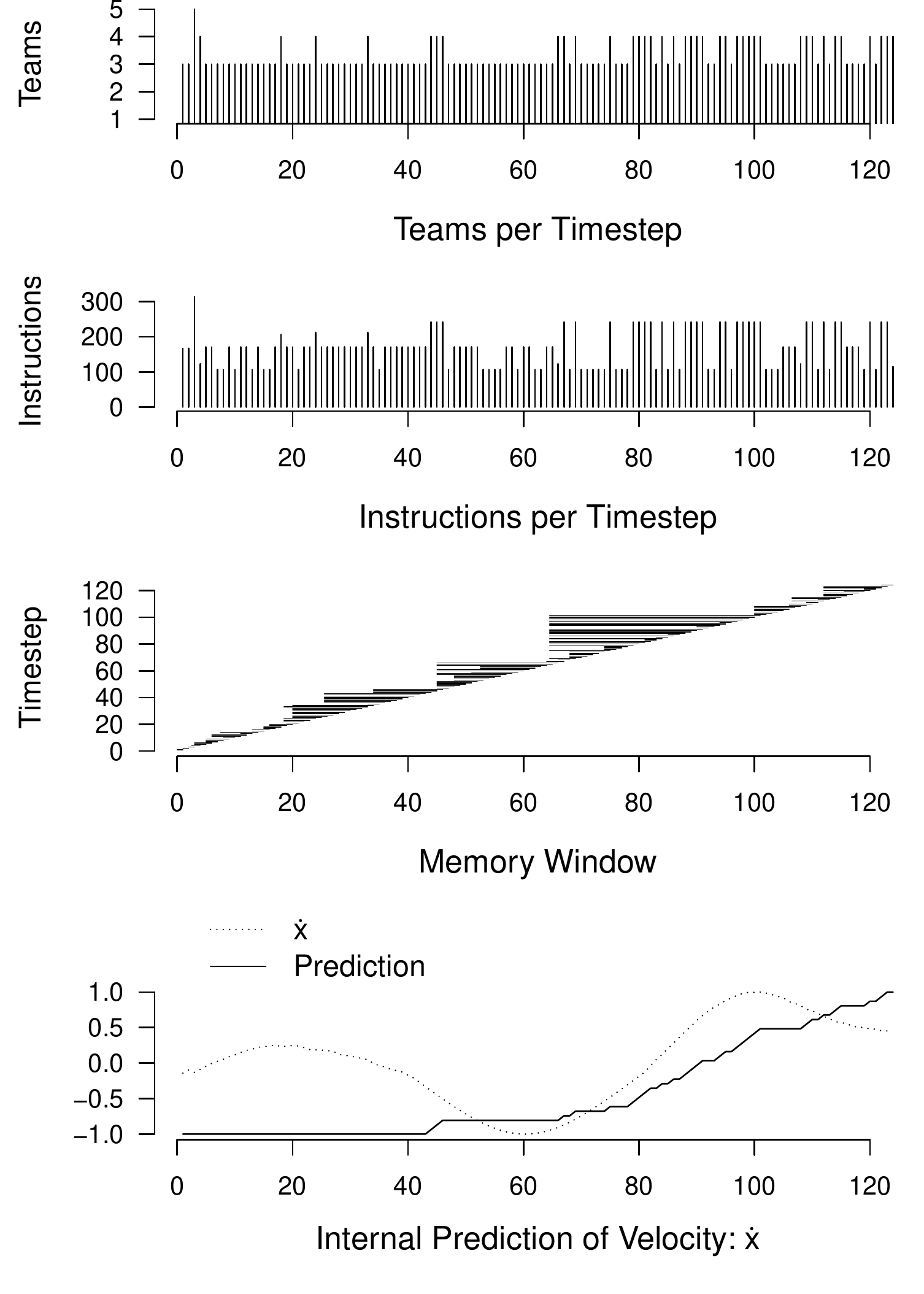}\label{fig:best_runTimeStats-tsk-6}}
	\caption{Time series data recorded during replay of best multi-task program graph (Figure \ref{fig:mtrl-graph}) in 1 episode of each task. x-axis is timesteps. See Sections \ref{sec:rtc}, \ref{sec:dynmem}, and \ref{sec:vel-prediction} for details. }
\label{fig:rtc-2}
\end{figure}

Figures \ref{fig:rtc-1} and  \ref{fig:rtc-2} show the run-time dynamics of the best multi-task program graph during 1 test episode in each task. Each node in the graph (Figure \ref{fig:mtrl-graph}) represents one team of programs. Every execution of the program graph begins at the root node and follows one path, which may terminate at any node. Furthermore, each team executes a unique subset of programs, each with a variable length list of instructions. Since the path of execution is dynamically selected, the computational complexity of program graph execution is also a dynamic property. The top two plots in Figures \ref{fig:best_runTimeStats-tsk-1} through \ref{fig:best_runTimeStats-tsk-6} show the run-time complexity for the champion program graph in each task. For example, the top plot in Figure \ref{fig:best_runTimeStats-tsk-4} indicates that the champion program graph executes between 2 and 6 teams per timestep in the pendulum environment. The rate of path switching fluctuates until timestep $\approx125$ and then stabilizes at 2 teams per timestep. This correlates with the 2 modes of behaviour required for pendulum: the agent must first rock the pendulum back and forth to gain enough momentum to swing the pendulum up to a vertical position (timestep 1 to $\approx125$). Then, a new mode of behaviour is required to balance the pendulum upright for the remainder of the episode. An animated example of this behaviour can be seen here: \href{https://vimeo.com/547319912}{Pendulum}. Dynamic run-time complexity improves the efficiency of model deployment when averaged over many timesteps. This is especially significant as complex (temporal) problems call for increasingly complex models. The most complex decision paths in any task execute 6 teams and roughly 300 instructions. This can be roughly compared with the D4PG deep neural network that holds several of the highest leaderboard scores, Section \ref{sec:leaderboard}. The D4PG agent network has two fully connect hidden layers with 400 and 300 neurons respectively. This implies that computing the forward pass at each timestep requires at least $400 \times 300 = 120,000$ calculations. While this can be computed in parallel on a \ac{GPU}, the relatively simple TPG agents do not require specialized hardware, making them suitable for operation on common embedded platforms such as the Raspberri Pi \cite{desnos21}. Note that the number of instructions per prediction in this work is significantly lower than that of our initial study in time series prediction \cite{kelly20b}. In this work, teams and programs are initialized with a much smaller size and mutation operators are slightly biased toward changes which result in simpler agents (See $progSize_{init}$, $p_{md,ma}$, $p_{delete,add}$, and $p_{atomic}$ in Table \ref{tbl:param}). 

\subsection{Dynamic Memory Access}\label{sec:dynmem}
Since program graphs are not provided with temporal state information (velocity), each program graph must define a mechanism for encoding observations within stateful memory registers, and recalling or resetting/overwriting these \text{memories} as required. Essentially, each program graph defines an internal encoding of the system state that is able to capture the temporal characteristics of \textit{any} task observed during training. Recall from Section \ref{sec:tpg} that each execution requires traversing one path through the program graph, where each team along the path will read/write to a unique set of stateful memory registers. As the active path changes over time, the agent's encoding of state also becomes dynamic. In particular, the "age" of memories accessed at any point in time effectively defines a memory window that fluctuates in width over time. The time point at which stateful memory registers are reset or left to accumulate is selected based on the current input as well as the content of stateful memory. Memory Window plots in Figures \ref{fig:rtc-1} and \ref{fig:rtc-2} depict the width of these dynamic memory windows at each timestep during test. The memory windows for time $t_1$ to $t_{n}$ are stacked vertically along the y-axis. Each horizontal line depicts the window width from the newest memory accessed (right-hand-side) to the oldest memory accessed (left-hand-side) at each timestep. Notice how the multi-task agent exhibits a unique pattern of dynamic memory access for each task.

\subsection{Internal Prediction of Unobservable State}\label{sec:vel-prediction}
In partially-observable \ac{MTRL}, dynamic memory access is critical for successful prediction of (unobservable) temporal properties of the system state. For example, in this work the agent is blind to system velocities. In order to select the best action at each timestep, the agent must \textit{predict} the velocities internally. Velocity at time $t+1$ can only be computed as a function of at least two observations made at previous timesteps and stored in memory. The Memory Window plots in Figures \ref{fig:rtc-1} and \ref{fig:rtc-2} show the maximum timespan from which these variables are drawn at each step. For example, the memory window for the pendulum task (Figure \ref{fig:best_runTimeStats-tsk-4}) fluctuates in size during the first mode of behaviour up to timestep $\approx 125$. These window-size fluctuations correlate to different paths through the graph being activated during this period. During this mode of behaviour, the pendulum is swinging back and forth and its angular velocity is sweeping through its entire range from positive to negative (see \href{https://vimeo.com/547319912}{Pendulum} animation). The agent is continuously using temporal memory to predict the pendulum's velocity internally, which is required information in order to produce actions (joint torques) that build the proper momentum to swing the pendulum up to vertical. We can confirm that the agent is actually constructing an internal model of velocity through this simple 2-step process:
\begin{enumerate}
    \item During replay, record the system velocity as well as the value stored in each memory register at each timestep. The best agent in this case contains 216 stateful registers (See Figure \ref{fig:mem-reg}) and the pendulum task has 1 unobservable velocity state variable ($\dot{\theta}$), giving us 217 time series recordings.
    \item Calculate Pearson correlation coefficient between the system velocity and all time series from agent memory, then identify the individual register that most strongly correlates with the system velocity. 
\end{enumerate}

The results of this analysis during replay in each task are plotted as Velocity Predictions in Figures \ref{fig:rtc-1} and \ref{fig:rtc-2}, where velocities and register time series are normalized in [-1,1]. Clearly, this agent is able to compute a useful internal prediction of system velocities while interacting with each task. The specific register containing the most correlated velocity predictions in Figures \ref{fig:best_runTimeStats-tsk-1} through \ref{fig:best_runTimeStats-tsk-6} are marked in Figure \ref{fig:mem-reg}. Note that the exact same register is used to store the velocity prediction for $\dot \theta_1$ in Acrobot and  $\dot x$ in CartCentering.

In the case of Pendulum, the internal prediction of velocity is very accurate during the first mode of behaviour up to timestep $\approx125$. Once the pendulum is vertically stabilized with an angular velocity near zero, memory-prediction is less critical because the agent can simply observe the pendulum's angle and apply a bang-bang force to keep it vertical. This behaviour can be seen in cell 6 of Figure \ref{fig:pend-decomp}. The pendulum is vertical at $cos(\theta)=1$. Blue and pink dots in this region indicate the agent is applying a positive/negative bang-bang force to keep the pendulum's angular velocity ($\dot \theta$) near zero (See \href{https://vimeo.com/547319912}{Pendulum} animation).

The ability to automatically define multiple memory windows with unique time delays and dynamically switch between them at run-time is critical in non-stationary and multi-task environments. Here, the agent exhibits unique patterns of dynamic memory access for tasks that have unique temporal properties and time constants (e.g. compare the rate of velocity change for CartPole and Pendulum in Figures \ref{fig:rtc-1} and \ref{fig:rtc-2}). Related studies have evolved "observation windows" in non-stationary time series forecasting, but still required human intuition in order to parameterize the window behaviour \cite{wagner07}. By contrast, our approach is entirely emergent.

%

\section{Conclusions and Future Work}\label{sec:conclusion}

TPG has been extended to support a modular temporal memory mechanism and simultaneous discrete and continuous output. We validate the new algorithm in a challenging multi-task reinforcement learning problem for which previous versions of TPG were not applicable. A single TPG agent can recognize and solve partially-observable versions of 6 \ac{RL} benchmark environments with a quality of behaviour that is competitive with the leading single-task, fully-observable deep learning approach. 

Evolving memory-prediction machines addresses all the challenges of \ac{MTRL} introduced in Section \ref{sec:mtrl}. Hierarchical program graphs built through compositional evolution support multi-task environments through automatic, hierarchical problem decomposition.  In short, agents can recombine multiple previously-independent generalist and specialist behaviours, and dynamically switch between then at run-time. This allows an agent to exploit positive inter-task transfer when tasks are related, and avoid negative transfer between disjoint tasks that require specialized behaviours. A multi-task selection process maintains a niche for generalist agents relative to each combination of tasks, ensuring useful hierarchical building blocks are always present in the population. A temporal memory mechanism allows agents to construct a dynamic internal world-model, which enables operation in partially-observable environments. Scalability is addressed by initializing the evolutionary search with simple programs and adapting their complexity entirely through environmental interaction. Variation operators are biased for simplicity, thus model complexity emerges gradually and is correlated with an increase in multi-task competence. The final complexity of a multi-task TPG agent is several orders of magnitude simpler than the leading deep learning agent trained from scratch for each task.

Compositional evolution with TPGs was initially demonstrated in high-dimensional (visual) \ac{MTRL} without any provision for temporal memory or support for mixed discrete and continuous actions spaces \cite{kelly18}. Given the developments presented herein, 
as well as recent progress made in multi-class image classification with TPG \cite{smith21}, we are interested to see how the approach operates in partially-observable visual \ac{RL} environments such as DeepMind Lab \cite{beattie2016}.  Future work will likely also address how the dynamic properties of TPG will behave in explicitly non-stationary time series environments \cite{wagner07,Agapitos2012} and dynamic memory tasks in which the input distribution changes significantly from training to test environments \cite{goyal2019recurrent}. The proposed temporal memory mechanism might also provide benefits under multi-task time series prediction, where the goal is to build a single model capable of forecasting multiple independent data streams \cite{Yang2019}. In short, this work significantly broadens the scope of our existing methods and opens a breadth of future research opportunities.

\begin{acknowledgements}
S.K. gratefully acknowledges support through the NSERC Postdoctoral Scholarship program. This material is based in part upon work supported by the National Science Foundation under Cooperative Agreement No. DBI-0939454 to the BEACON Center for Evolution in Action at Michigan State University. W.B. acknowledges support from the John R. Koza Endowment fund for part of this work. Michigan State University provided computational resources through the Institute for Cyber-Enabled Research. Additional support provided by ACENET, Calcul Québec, Compute Ontario and WestGrid, and Compute Canada (www.computecanada.ca). Any opinions, findings, and conclusions or recommendations expressed in this material are those of the author(s) and do not necessarily reflect the views of the National Science Foundation.
\end{acknowledgements}

%
 \section*{Conflict of interest}

 The authors declare that they have no conflict of interest.

\bibliographystyle{spmpsci}      
\bibliography{skelly-gpem2021}   

\begin{thebibliography}{10}
\providecommand{\url}[1]{{#1}}
\providecommand{\urlprefix}{URL }
\expandafter\ifx\csname urlstyle\endcsname\relax
  \providecommand{\doi}[1]{DOI~\discretionary{}{}{}#1}\else
  \providecommand{\doi}{DOI~\discretionary{}{}{}\begingroup
  \urlstyle{rm}\Url}\fi

\bibitem{Agapitos2012}
Agapitos, A., O'Neill, M., Brabazon, A.: Genetic programming for the induction
  of seasonal forecasts: A study on weather derivatives.
\newblock In: M.~Doumpos, C.~Zopounidis, P.M. Pardalos (eds.) Financial
  Decision Making Using Computational Intelligence, pp. 159--188. Springer US,
  Boston, MA (2012)

\bibitem{banino2020memo}
Banino, A., Badia, A.P., Koster, R., Chadwick, M.J., Zambaldi, V., Hassabis,
  D., Barry, C., Botvinick, M., Kumaran, D., Blundell, C.: Memo: A deep network
  for flexible combination of episodic memories.
\newblock arXiv \textbf{2001.10913} (2020)

\bibitem{barthmaron2018distributed}
Barth-Maron, G., Hoffman, M.W., Budden, D., Dabney, W., Horgan, D., TB, D.,
  Muldal, A., Heess, N., Lillicrap, T.: Distributed distributional
  deterministic policy gradients.
\newblock arXiv \textbf{1804.08617} (2018)

\bibitem{beattie2016}
Beattie, C., Leibo, J.Z., Teplyashin, D., Ward, T., Wainwright, M., Küttler,
  H., Lefrancq, A., Green, S., Valdés, V., Sadik, A., Schrittwieser, J.,
  Anderson, K., York, S., Cant, M., Cain, A., Bolton, A., Gaffney, S., King,
  H., Hassabis, D., Legg, S., Petersen, S.: {DeepMind Lab}.
\newblock arXiv \textbf{1612.03801} (2016)

\bibitem{brameier07}
Brameier, M., Banzhaf, W.: Linear Genetic Programming.
\newblock Springer (2007)

\bibitem{brockman2016openai}
Brockman, G., Cheung, V., Pettersson, L., Schneider, J., Schulman, J., Tang,
  J., Zaremba, W.: {OpenAI Gym}.
\newblock arXiv \textbf{1606.01540} (2016)

\bibitem{DEramo2020Sharing}
D'Eramo, C., Tateo, D., Bonarini, A., Restelli, M., Peters, J.: Sharing
  knowledge in multi-task deep reinforcement learning.
\newblock In: International Conference on Learning Representations (2020).
\newblock \urlprefix\url{https://openreview.net/forum?id=rkgpv2VFvr}

\bibitem{desnos21}
Desnos, K., Sourbier, N., Raumer, P.Y., Gesny, O., Pelcat, M.: {Gegelati:
  Lightweight Artificial Intelligence through Generic and Evolvable Tangled
  Program Graphs}.
\newblock In: Workshop on Design and Architectures for Signal and Image
  Processing (14th Edition), DASIP '21, p. 35–43. ACM, New York, NY, USA
  (2021).
\newblock \doi{10.1145/3441110.3441575}.
\newblock \urlprefix\url{https://doi.org/10.1145/3441110.3441575}

\bibitem{fernando2017}
Fernando, C., Banarse, D., Blundell, C., Zwols, Y., Ha, D., Rusu, A.A.,
  Pritzel, A., Wierstra, D.: Pathnet: Evolution channels gradient descent in
  super neural networks.
\newblock arXiv \textbf{1701.08734} (2017)

\bibitem{fu2019deep}
Fu, H., Tang, H., Hao, J., Lei, Z., Chen, Y., Fan, C.: Deep multi-agent
  reinforcement learning with discrete-continuous hybrid action spaces.
\newblock arXiv \textbf{1903.04959} (2019)

\bibitem{gomez05}
Gomez, F.J., Schmidhuber, J.: Co-evolving recurrent neurons learn deep memory
  pomdps.
\newblock In: Proceedings of the 7th Annual Conference on Genetic and
  Evolutionary Computation, GECCO '05, p. 491–498. ACM, New York, NY, USA
  (2005).
\newblock \doi{10.1145/1068009.1068092}.
\newblock \urlprefix\url{https://doi.org/10.1145/1068009.1068092}

\bibitem{goyal2019recurrent}
Goyal, A., Lamb, A., Hoffmann, J., Sodhani, S., Levine, S., Bengio, Y.,
  Sch\"olkopf, B.: Recurrent independent mechanisms.
\newblock arXiv \textbf{1909.10893} (2019)

\bibitem{Greff2017}
Greff, K., Srivastava, R.K., Koutník, J., Steunebrink, B.R., Schmidhuber, J.:
  Lstm: A search space odyssey.
\newblock IEEE Transactions on Neural Networks and Learning Systems
  \textbf{28}(10), 2222--2232 (2017).
\newblock \doi{10.1109/TNNLS.2016.2582924}

\bibitem{Hessel2019}
Hessel, M., Soyer, H., Espeholt, L., Czarnecki, W., Schmitt, S., van Hasselt,
  H.: Multi-task deep reinforcement learning with popart.
\newblock Proceedings of the AAAI Conference on Artificial Intelligence
  \textbf{33}(01), 3796--3803 (2019).
\newblock \doi{10.1609/aaai.v33i01.33013796}.
\newblock \urlprefix\url{https://ojs.aaai.org/index.php/AAAI/article/view/4266}

\bibitem{heywood_evolutionary_2015}
Heywood, M.I.: Evolutionary model building under streaming data for
  classification tasks: opportunities and challenges.
\newblock Genetic Programming and Evolvable Machines \textbf{16}(3), 283--326
  (2015)

\bibitem{Holland:1985}
Holland, J.H.: Properties of the bucket brigade algorithm.
\newblock Proceedings of an International Conference on Genetic Algorithms and
  their Applications pp. 1--7 (1985)

\bibitem{skellyphd}
Kelly, S.: Scaling genetic programming to challenging reinforcement tasks
  through emergent modularity.
\newblock Ph.D. thesis, Faculty of Computer Science, Dalhousie University
  (2018)

\bibitem{tpgSource}
Kelly, S.: TPG Source Code (2020).
\newblock Available at \url{http://stephenkelly.ca/?q=research}

\bibitem{acrobotA}
Kelly, S.: Acrobot Animation (2021).
\newblock Available at
  \href{https://vimeo.com/547319719}{https://vimeo.com/547319719}

\bibitem{cartCenteringA}
Kelly, S.: CartCentering Animation (2021).
\newblock Available at
  \href{https://vimeo.com/547319756}{https://vimeo.com/547319756}

\bibitem{cartPoleA}
Kelly, S.: CartPole Animation (2021).
\newblock Available at
  \href{https://vimeo.com/547319808}{https://vimeo.com/547319808}

\bibitem{mountainCarCA}
Kelly, S.: MountainCarC. Animation (2021).
\newblock Available at
  \href{https://vimeo.com/547319863}{https://vimeo.com/547319863}

\bibitem{pendulumA}
Kelly, S.: Pendulum Animation (2021).
\newblock Available at
  \href{https://vimeo.com/547319912}{https://vimeo.com/547319912}

\bibitem{kelly20a}
Kelly, S., Banzhaf, W.: Temporal memory sharing in visual reinforcement
  learning.
\newblock In: W.~Banzhaf, L.~Spector, L.~Sheneman (eds.) Genetic Programming
  Theory and Practice XVII, pp. 101--119. Springer International Publishing,
  Cham (2020)

\bibitem{kelly18b}
Kelly, S., Heywood, M.I.: {Discovering Agent Behaviors Through Code Reuse:
  Examples From Half-Field Offense and Ms. Pac-Man}.
\newblock IEEE Transactions on Games \textbf{10}(2), 195--208 (2018)

\bibitem{kelly18}
Kelly, S., Heywood, M.I.: Emergent solutions to high-dimensional multitask
  reinforcement learning.
\newblock Evolutionary Computation \textbf{26}(3), 347--380 (2018)

\bibitem{kelly20b}
Kelly, S., Newsted, J., Banzhaf, W., Gondro, C.: A modular memory framework for
  time series prediction.
\newblock In: Proceedings of the 2020 Genetic and Evolutionary Computation
  Conference, GECCO '20, p. 949–957. ACM, New York, NY, USA (2020).
\newblock \doi{10.1145/3377930.3390216}.
\newblock \urlprefix\url{https://doi.org/10.1145/3377930.3390216}

\bibitem{kingman_1978}
Kingman, J.F.C.: A simple model for the balance between selection and mutation.
\newblock Journal of Applied Probability \textbf{15}(1), 1--12 (1978)

\bibitem{Kirkpatrick3521}
Kirkpatrick, J., Pascanu, R., Rabinowitz, N., Veness, J., Desjardins, G., Rusu,
  A.A., Milan, K., Quan, J., Ramalho, T., Grabska-Barwinska, A., Hassabis, D.,
  Clopath, C., Kumaran, D., Hadsell, R.: Overcoming catastrophic forgetting in
  neural networks.
\newblock Proceedings of the National Academy of Sciences \textbf{114}(13),
  3521--3526 (2017)

\bibitem{Koza1992}
Koza, J.R.: Genetic Programming: On the Programming of Computers by Means of
  Natural Selection.
\newblock MIT Press (1992)

\bibitem{lillicrap2015continuous}
Lillicrap, T.P., Hunt, J.J., Pritzel, A., Heess, N., Erez, T., Tassa, Y.,
  Silver, D., Wierstra, D.: Continuous control with deep reinforcement
  learning.
\newblock arXiv \textbf{1509.02971} (2015)

\bibitem{metz2017discrete}
Metz, L., Ibarz, J., Jaitly, N., Davidson, J.: {Discrete Sequential Prediction
  of Continuous Actions for Deep RL}.
\newblock arXiv \textbf{1705.05035} (2017)

\bibitem{mnih15}
Mnih, V., Kavukcuoglu, K., Silver, D., Rusu, A.A., Veness, J., Bellemare, M.G.,
  Graves, A., Riedmiller, M., Fidjeland, A.K., Ostrovski, G., Petersen, S.,
  Beattie, C., Sadik, A., Antonoglou, I., King, H., Kumaran, D., Wierstra, D.,
  Legg, S., Hassabis, D.: Human-level control through deep reinforcement
  learning.
\newblock Nature \textbf{518}(7540), 529--533 (2015)

\bibitem{Nedelcu02}
Nedelcu, A.M., Michod, R.E.: Evolvability, modularity, and individuality during
  the transition to multicellularity in volvocalean green algae.
\newblock In: G.~Schlosser, G.~Wagner (eds.) Modularity in Development and
  Evolution, pp. 470--489. Chicago Press (2002)

\bibitem{Neftci2019}
Neftci, E.O., Averbeck, B.B.: Reinforcement learning in artificial and
  biological systems.
\newblock Nature Machine Intelligence \textbf{1}(3), 133--143 (2019).
\newblock \doi{10.1038/s42256-019-0025-4}.
\newblock \urlprefix\url{https://doi.org/10.1038/s42256-019-0025-4}

\bibitem{oh2016control}
Oh, J., Chockalingam, V., Singh, S., Lee, H.: {Control of Memory, Active
  Perception, and Action in Minecraft}.
\newblock arXiv \textbf{1605.09128} (2016)

\bibitem{EVCO_a_00080}
Preen, R.J., Bull, L.: {Dynamical Genetic Programming in Xcsf}.
\newblock Evolutionary Computation \textbf{21}(3), 361--387 (2013)

\bibitem{annurev-control}
Recht, B.: A tour of reinforcement learning: The view from continuous control.
\newblock Annual Review of Control, Robotics, and Autonomous Systems
  \textbf{2}(1), 253--279 (2019)

\bibitem{rusu2016policy}
Rusu, A.A., Colmenarejo, S.G., Gulcehre, C., Desjardins, G., Kirkpatrick, J.,
  Pascanu, R., Mnih, V., Kavukcuoglu, K., Hadsell, R.: Policy distillation.
\newblock arXiv \textbf{1511.06295} (2016)

\bibitem{rusu2016}
Rusu, A.A., Rabinowitz, N.C., Desjardins, G., Soyer, H., Kirkpatrick, J.,
  Kavukcuoglu, K., Pascanu, R., Hadsell, R.: Progressive neural networks.
\newblock arXiv \textbf{1606.04671} (2016)

\bibitem{Simon62}
Simon, H.A.: The architecture of complexity.
\newblock Proceedings of the American Philosophical Society \textbf{106},
  467--482 (1962)

\bibitem{smith21}
Smith, R.J., Amaral, R., Heywood, M.I.: {Evolving Simple Solutions to the
  CIFAR-10 Benchmark using Tangled Program Graphs}.
\newblock In: Proceedings of the 2021 IEEE Congress of Evolutionary Computation
  (CEC), p. to appear (2021)

\bibitem{smith19b}
Smith, R.J., Heywood, M.I.: {Evolving Dota 2 Shadow Fiend Bots Using Genetic
  Programming with External Memory}.
\newblock In: Proceedings of the Genetic and Evolutionary Computation
  Conference, GECCO '19, pp. 179--187. ACM, New York, NY, USA (2019)

\bibitem{smith19a}
Smith, R.J., Heywood, M.I.: A model of external memory for navigation in
  partially observable visual reinforcement learning tasks.
\newblock In: L.~Sekanina, T.~Hu, N.~Louren{\c{c}}o, H.~Richter,
  P.~Garc{\'i}a-S{\'a}nchez (eds.) Genetic Programming, pp. 162--177. Springer
  International Publishing, Cham (2019)

\bibitem{sutton98}
Sutton, R.R., Barto, A.G.: Reinforcement Learning: An Introduction.
\newblock MIT Press (1998)

\bibitem{electronics9091363}
Vithayathil~Varghese, N., Mahmoud, Q.H.: A survey of multi-task deep
  reinforcement learning.
\newblock Electronics \textbf{9}(9) (2020).
\newblock \doi{10.3390/electronics9091363}.
\newblock \urlprefix\url{https://www.mdpi.com/2079-9292/9/9/1363}

\bibitem{Wagner96}
Wagner, G.P., Altenberg, L.: Perspective: Complex adaptations and the evolution
  of evolvability.
\newblock Evolution \textbf{50}(3), 967--976 (1996)

\bibitem{wagner07}
{Wagner}, N., {Michalewicz}, Z., {Khouja}, M., {McGregor}, R.R.: {Time Series
  Forecasting for Dynamic Environments: The DyFor Genetic Program Model}.
\newblock IEEE Transactions on Evolutionary Computation \textbf{11}(4),
  433--452 (2007)

\bibitem{watson_modular_2005}
Watson, R.A., Pollack, J.B.: Modular interdependency in complex dynamical
  systems.
\newblock Artificial Life \textbf{11}(4), 445--457 (2005)

\bibitem{Yang01}
Yang, A.S.: Modularity, evolvability, and adaptive radiations: a comparison of
  the hemi- and holometabolous insects.
\newblock Evolution and Development \textbf{3}(2), 59--72 (2001)

\bibitem{Yang2019}
Yang, M., Hu, Q., Wang, Y.: Multi-task learning method for hierarchical time
  series forecasting.
\newblock In: I.V. Tetko, V.~K{\r{u}}rkov{\'a}, P.~Karpov, F.~Theis (eds.)
  Artificial Neural Networks and Machine Learning -- ICANN 2019: Text and Time
  Series, pp. 474--485. Springer International Publishing, Cham (2019)

\bibitem{yang2020multitask}
Yang, R., Xu, H., Wu, Y., Wang, X.: Multi-task reinforcement learning with soft
  modularization.
\newblock arXiv \textbf{2003.13661} (2020)

\bibitem{yannakakis2018}
Yannakakis, G.N., Togelius, J.: {Artificial Intelligence and Games}.
\newblock Springer (2018).
\newblock \url{http://gameaibook.org}

\end{thebibliography}

\end{document}